\pdfoutput=1
\documentclass{article}

\newif\ifincludechecklist
\includechecklistfalse

\PassOptionsToPackage{numbers,compress}{natbib}

 \usepackage[main, final]{neurips_2026}

\usepackage[dvipsnames]{xcolor}         
\usepackage[colorlinks=true]{hyperref}
\definecolor{linkscolor}{RGB}{54,117,136}
\hypersetup{
  colorlinks=true,
  allcolors=linkscolor
}
\usepackage{url}            
\usepackage{booktabs}       
\usepackage{amsfonts}       
\usepackage{nicefrac}       
\usepackage{microtype}      
\usepackage{amsmath}
\usepackage{amssymb}
\usepackage{mathtools}
\usepackage{amsthm}
\usepackage{bbm}
\usepackage{inconsolata}
\usepackage{graphicx}
\usepackage{multicol}
\usepackage{multirow}
\usepackage{subcaption}
\usepackage{wrapfig}
\usepackage{booktabs}
\usepackage{colortbl}
\usepackage{cleveref}
\usepackage[most]{tcolorbox}
\usepackage{thmtools}

\newtcolorbox[
auto counter,
crefname={RQ}{RQ}
]{rqbox}[2][]{%
colframe=black!40,
colback=gray!0,
colbacktitle=gray!40,
fonttitle=\bfseries,
arc=2pt,outer arc=2pt,
enhanced,
attach boxed title to top left={yshift=-1pt},
boxed title style={arc=0pt,outer arc=0pt},
#1
}

\newcommand{\entities}{\mathcal{E}}
\newcommand{\relations}{\mathcal{R}}
\newcommand{\edges}{\mathcal{T}}
\newcommand{\edgestrain}{\mathcal{T}^{\text{train}}}
\newcommand{\edgesvalid}{\mathcal{T}^{\text{valid}}}
\newcommand{\edgestest}{\mathcal{T}^{\text{test}}}
\newcommand{\graph}{\mathcal{G}}
\newcommand{\logit}{\operatorname{logit}}
\newcommand{\cosim}{\operatorname{sim}}
\newcommand{\ncosim}{\overline{\operatorname{sim}}}
\newcommand{\prefvar}{P_s^{\odot}}
\newcommand{\prefpos}{P_s^{+}}
\newcommand{\prefneg}{P_s^{-}}
\newcommand{\query}{q(v_1, \ldots, v_N)}

\let\mc\multicolumn

\definecolor{sagegreen}{HTML}{85AA69}
\definecolor{mutedred}{HTML}{BF616A}
\newcommand{\promoted}{\textcolor{sagegreen}{\small$\blacktriangle$}}
\newcommand{\demoted}{\textcolor{mutedred}{\small$\blacktriangledown$}}

\title{Similarity-Constrained Reweighting for\\ Complex Query Answering on Knowledge Graphs}

\author{%
Daniel Daza$^{1,2,7}$ \quad
Alberto Bernardi$^{3}$ \quad
Luca Costabello$^{3}$ \quad
Christophe Gueret$^{3}$ \\
\textbf{Michael Cochez}$^{4,5,6}$ \quad
\textbf{Martijn Schut}$^{7}$ \\[0.5em]
$^{1}$Vrije Universiteit Amsterdam \\
$^{2}$Eindhoven University of Technology \\
$^{3}$Accenture Labs \\
$^{4}$ELLIS Institute Finland \\
$^{5}$\AA bo Akademi University, Turku, Finland \\
$^{6}$Elsevier Discovery Lab, Amsterdam \\
$^{7}$Translational AI Laboratory, Department of Laboratory Medicine\\
Amsterdam University Medical Center, Vrije Universiteit Amsterdam \\[0.5em]
\texttt{d.dazacruz@vu.nl} \quad
\texttt{alberto.bernardi@accenture.com} \\
\texttt{luca.costabello@accenture.com}\quad
\texttt{christophe.gueret@accenture.com} \\
\texttt{michael.cochez@abo.fi} \quad
\texttt{m.c.schut@amsterdamumc.nl}
}

\begin{document}

\maketitle

\begin{abstract}
Machine learning models for Complex Query Answering (CQA) on knowledge graphs estimate the likelihood of answers that are not reachable via direct traversal. Recent work has proposed extending such queries with \emph{soft entity constraints}, which require the \emph{target variable} to be similar or dissimilar to specified sets of entities. We study a more general formulation where similarity constraints are specified on an arbitrary variable in a query, and introduce SCORE: a computationally efficient and interpretable method for incorporating similarity constraints on an arbitrary variable of a query. SCORE is based on a lightweight and interpretable score adjustment function that only requires tuning two hyperparameters on a validation set. Our experiments show that SCORE effectively incorporates preferences while maintaining strong query answering performance. It significantly improves ranking quality when constraints apply to the target variable and performs effectively in the more general case involving constraints on intermediate variables, and its performance remains robust when applied to several existing methods for CQA.

Our implementation is available at \url{https://github.com/dfdazac/score}.
\end{abstract}

\section{Introduction}

Knowledge graphs (KGs) have emerged as a fundamental data structure for organizing and reasoning about real-world information, encoding millions of entities and their relationships across diverse domains including encyclopedic knowledge~\citep{bollacker_freebase_2008}, biomedical data~\citep{himmelstein_systematic_2017}, and scientific literature~\citep{ji_survey_2020}. Their structured representation enables reasoning tasks that go beyond simple fact retrieval, supporting complex queries that involve multiple logical operations such as conjunction, disjunction, and existential quantification. These capabilities have made KGs relevant in applications such as question answering and recommendation~\citep{zhou2020interactive}, drug discovery~\citep{daza2025bioblp}, and knowledge-based reasoning~\citep{hogan_knowledge_2021}.

However, knowledge graphs are inherently incomplete, missing a substantial fraction of true facts~\citep{nickel_review_2015}. This incompleteness poses significant challenges for traditional symbolic query processing, which can only retrieve answers that are explicitly present in the graph. To address this limitation, machine learning approaches for \emph{complex query answering} have been developed~\citep{hamilton2018gqe,ren_query2box_2020,ren_beta_2020,ren2024neural}, which learn to estimate the likelihood of potential answers by reasoning over incomplete knowledge graphs.

Despite these advances, current complex query answering systems are limited to expressing queries through first-order logic, which restricts their ability to capture nuanced preferences and domain-specific requirements. Recent work has begun to address this limitation by introducing \emph{soft entity constraints}~\citep{daza2025interactive}, which allows specifying that query answers should be similar or dissimilar to given sets of entities. For instance, when querying for ``drugs treating diseases that have headache as a symptom,'' a user might specify preferences such as ``drugs similar to Ibuprofen''. While this formulation represents a significant step towards more expressive interfaces for querying a KG, it only supports similarity constraints on the final answers being sought, also known as the \emph{target variable}. This limitation limits expressiveness, as constraints can also involve intermediate variables that appear during query processing. For example, one might specify constraints not only on the drug (target variable) but also on the diseases (the intermediate variable), such as ``diseases like migraine''.

\begin{figure*}[t]
    \centering
    \includegraphics[width=\linewidth]{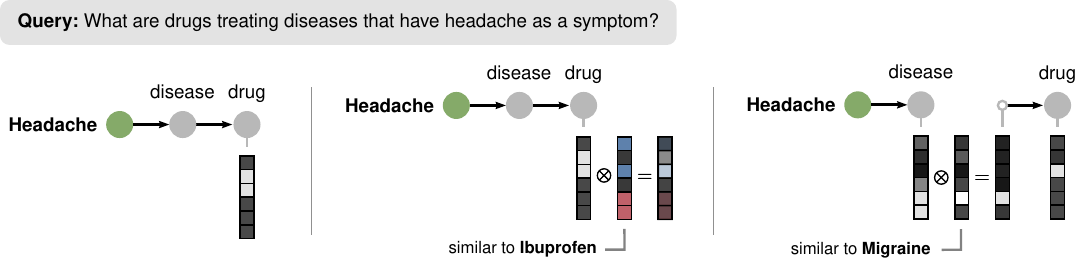}
    \caption{Illustration of three approaches for query answering on knowledge graphs, with vectors indicating entity scores at different steps. \textbf{Left}: a traditional query with purely logical constraints. \textbf{Middle}: incorporating similarity constraints in NQR~\citep{daza2025interactive}, which supports only the target variable via uncalibrated score adjustments (shown in red and blue). \textbf{Right}: SCORE, which supports similarity constraints on arbitrary variables through normalized similarity scores.}
    \label{fig:simcqa_cases}
\end{figure*}

When constraints apply only to target variables, similarity-based score adjustments can be applied as a post-processing step after the query has been evaluated by a base Complex Query Answering (CQA) model. Incorporating constraints on intermediate variables, on the other hand, is a more challenging problem, as they must be \emph{integrated} into the query processing itself. Crucially, this integration depends on operating with \emph{normalized scores}, rather than uncalibrated similarity scores, so that they can be meaningfully combined with other probabilistic components of the query answering model. Such an integration must preserve the logical structure of the query while incorporating similarity constraints at arbitrary positions, which is not supported by existing methods ~\citep{daza2025interactive}.

In this work, we address this fundamental limitation by studying the problem of \emph{Similarity-constrained Complex Query Answering} (SimCQA) in its full generality, where similarity constraints can be specified on \textbf{any variable} within a complex query. We then propose SCORE (\textbf{S}imilarity-\textbf{CO}nstrained \textbf{RE}weighting), a computationally efficient method that extends a base CQA model that traverses the query graph by propagating entity score vectors.

The key innovation of SCORE lies in its lightweight and interpretable score adjustment mechanism, which uses pre-trained entity embeddings to compute normalized similarity scores and integrates them into the query processing pipeline, as illustrated in~\Cref{fig:simcqa_cases} on the right. Unlike approaches that require expensive end-to-end neural optimization, SCORE introduces only two hyperparameters that can be efficiently tuned on a validation set and leaves the base CQA model unchanged. Our contributions are summarized as follows:

\begin{itemize}
\item We formalize the problem of similarity-constrained complex query answering with constraints on arbitrary variables, generalizing previous work.
\item We introduce SCORE, a lightweight and interpretable method for incorporating similarity constraints at any position in complex queries, requiring minimal computational overhead and hyperparameter tuning.
\item We conduct comprehensive experiments on three challenging knowledge graph benchmarks (FB15K-237, NELL-995, and Hetionet), demonstrating that SCORE achieves substantial improvements in ranking quality and constraint satisfaction that are maintained across different base CQA models.
\end{itemize}

\section{Related Work}
\label{sec:related_work}

\paragraph{Complex query answering.}
Neural approaches to complex query answering (CQA) have evolved from simple link prediction to multi-hop logical queries~\cite{ren2024neural}. Early knowledge graph embedding methods focused on predicting individual missing edges~\citep{nickel2011rescal,bordes2013transe,trouillon2016complex,lacroix2018complex_n3,sun2019rotate}, learning dense representations that capture relational patterns in the graph structure. The field has since progressed to handle complex queries involving multiple entities, relations, and logical operations such as conjunction, disjunction, and existential quantification. Modern \emph{neuro-symbolic} approaches can be broadly categorized based on their query expressiveness: methods supporting smaller subsets of first-order logic~\citep{hamilton2018gqe,daza2020message,ren_query2box_2020,ren_beta_2020,arakelyan2021cqd,arakelyan2023cqdA,bai2023qto,zhu_neural-symbolic_2022} and those handling more general graph pattern queries~\citep{cucumides_unravl_2024,yin_rethinking_2024}.

A fundamental limitation shared by all these approaches is their reliance on \emph{hard logical constraints}: queries can only express conditions that are precisely definable in first-order logic. This restriction prevents users from incorporating nuanced preferences or domain-specific similarity requirements that are naturally expressed through soft constraints. Our work addresses this gap by extending neural query answering to support similarity-based preferences while preserving the logical structure and efficiency of existing methods. LLM- and tool-based graph traversal methods~\citep{jinhao2023structgpt,sun2024think} address a complementary setting where natural-language questions guide retrieved subgraph exploration, which also requires access to textual descriptions. Our work focuses on graph-structured queries and does not assume access to textual descriptions.

\paragraph{Similarity constraints for complex queries.}
The first work to extend logical query answering with similarity constraints was introduced by~\citet{daza2025interactive}, which augmented first-order logic queries with \emph{soft entity constraints} expressing similarity or dissimilarity preferences for query answers. This enriched CQA with preferences that cannot be captured through logical operators alone. However, this prior setting is limited to constraints on the \emph{target variable}, where similarity can be incorporated after query execution as a reranking step. Constraints on intermediate variables create a different computational problem: preference information must enter the query execution process and then be propagated through subsequent projections, intersections, and negations. Uncalibrated post-hoc score shifts are therefore insufficient, because intermediate scores must remain compatible with the fuzzy operators used by the base CQA model. In contrast, SCORE maps similarities into normalized score updates and injects them into intermediate fuzzy vectors, enabling similarity constraints to influence downstream reasoning while preserving the query structure.

\section{Preliminaries}

\paragraph{Knowledge graphs.} 
We define a knowledge graph as a tuple $\graph = (\entities, \relations, \edges)$, where $\entities$ is the set of entities, $\relations$ is the set of binary relations, and $\edges$ is the set of triples $(h, r, t)$ with $h, t \in \entities$ and $r \in \relations$.

\paragraph{Complex queries.} 
Queries over KGs are expressed as first-order logic formulas $q(v_1, \ldots, v_N)$ with free variables $v_1, \ldots, v_N$ ranging over $\entities$. Formulas are built from atoms over relations in $\relations$ (and their negations), constants denoting specific entities in $\entities$, and logical connectives such as conjunction ($\wedge$) and disjunction ($\vee$).
For example, consider the question  
\emph{“What drugs treat diseases that have headache as a symptom?”}.  
The corresponding query for this question is the following:
\begin{equation}
    q(v_1,v_2) =\; \texttt{SymptomOf}(\texttt{Headache}, v_1) \wedge\; \texttt{TreatedBy}(v_1, v_2).
    \label{eq:query_example}
\end{equation}

The free variables in a query can be assigned to specific entities in $\entities$. For an assignment $e_1, \ldots, e_N \in \entities$, we write $q(e_1, \ldots, e_N)$ for the resulting \textbf{grounded statement}. We write $\graph \models q(e_1, \ldots, e_N)$ if the set of triples $\edges$ in the KG entails this grounded formula.

Continuing the example, suppose $e_1 = \texttt{Migraine}$ and $e_2 = \texttt{Ibuprofen}$. We then obtain the grounded statement:
\begin{equation}
    \texttt{SymptomOf}(\texttt{Headache}, \texttt{Migraine}) \wedge\; \texttt{TreatedBy}(\texttt{Migraine}, \texttt{Ibuprofen}).
\end{equation}
If both atoms appear in $\edges$, then $\graph \models q(\texttt{Migraine}, \texttt{Ibuprofen})$.

\paragraph{Answer set.} 
Given a query formula $q(v_1, \ldots, v_N)$, one may choose a free variable $v_i$ as a designated \emph{target variable}. Denoting as $\boldsymbol{v}_{\neg i}$ all free variables except $v_i$, the answer set $A_i(q)$ is defined as the set of entities that can be assigned to $v_i$ such that the query holds in the graph for some assignment of the remaining variables:
\begin{equation}
    A_i(q) = \lbrace e\in\entities \mid \exists\boldsymbol{v}_{\neg i} \in \entities : \graph\models q(\boldsymbol{v}_{\neg i}, v_i = e) \rbrace.
\label{eq:answer_set}
\end{equation}

Returning to the earlier query in~\Cref{eq:query_example}, suppose we choose \(v_2\) (the drug) as the target variable. 
Then the answer set is
\begin{equation}
    A_2(q) = \{\, e \in \entities \mid \exists v_1 \in \entities : \graph \models \texttt{SymptomOf}(\texttt{Headache}, v_1) \land \texttt{TreatedBy}(v_1, e) \,\}.
\end{equation}

These definitions form the basis of prior work on complex query answering (CQA), where the goal is to predict answer sets correctly when the graph is incomplete (requiring accurate estimation of the likelihood of entailments $\graph \models q(a_1, \ldots, a_N)$)~\citep{hamilton2018gqe,daza2020message,ren_query2box_2020,ren_beta_2020}. They are also adopted by~\citet{daza2025interactive} to introduce similarity constraints over a single target variable.

\section{Similarity-constrained Complex Query Answering}

We now consider the more general problem where similarity constraints are specified over an arbitrary variable in a query.

\paragraph{Similarity constraints.} A similarity constraint is intended to capture notions such as ``diseases like migraine''. Formally, a similarity constraint is a boolean predicate $s$ such that for an entity $e\in\entities$, $s(e)$ is true if $e$ meets the similarity constraint, and false otherwise.

\paragraph{Constrained answer set.} Similarity constraints allow modifying the answer set (\Cref{eq:answer_set}) by additionally restricting certain variables to meet the constraint. Given a similarity constraint $s$ applied to variable $v_j$ in a query $\query$, we denote the \emph{constrained} answer set as follows:
\begin{equation}
    \hat{A}_i(q, s, j) = \lbrace e\in\entities \mid \exists\boldsymbol{v}_{\neg i} : (\graph\models q(\boldsymbol{v}_{\neg i}, v_i = e)) \wedge s(v_j) \rbrace.
\label{eq:constrained_answer_set}
\end{equation}
This can be read as the set of entities $e$ that may serve as answers for $v_i$, for which there exists an assignment of the remaining variables such that the query holds in the KG, \emph{and} the assignment to $v_j$ satisfies the similarity constraint $s$.

Because similarity constraints may vary across queries and domain context, $s$ is not fully observed. Instead, we assume access to a small set of $k$ (in the order of tens) labeled examples in a \emph{preference set}:
\begin{equation}
    P_s = \lbrace(e_1, s(e_1)), \ldots, (e_k, s(e_k))\rbrace.
\label{eq:preference_set}
\end{equation}

Suppose the similarity constraint expresses the notion of \emph{“diseases like migraine”}. Then an example preference set is $P_s = \{(\texttt{Migraine}, \text{true}), (\texttt{ClusterHeadache}, \text{true}), (\texttt{Asthma}, \text{false})\}$.
Together, these definitions provide the foundation for the problem we study in this work.

\begin{rqbox}

    \paragraph{Similarity-constrained Complex Query Answering.} 
Given a knowledge graph $\graph$, a query $q(v_1, \ldots, v_N)$ with target variable $v_i$, and a similarity constraint $s$ applied to variable $v_j$, the task of Similarity-constrained Complex Query Answering (SimCQA) consists of computing the constrained answer set $\hat{A}_i(q, s, j)$ (\Cref{eq:constrained_answer_set}), where the similarity constraint $s$ is estimated from a preference set $P_s$ (\Cref{eq:preference_set}).
\end{rqbox}

This formulation subsumes prior work. If we set $v_i=v_j$, we obtain the setting of~\citet{daza2025interactive} where similarity constraints are applied to a single target variable. If, in addition, we define a similarity function $s(e)=\text{true}$ for all $e \in \entities$, we recover standard CQA with purely logical queries.

\section{Similarity-Constrained Reweighting}

\begin{figure}[t]
    \centering
    \includegraphics[width=0.7\linewidth]{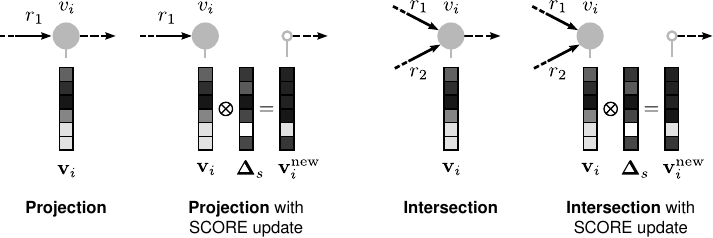}
    \caption{Existing CQA methods compute fuzzy vectors via \emph{projection} and \emph{intersection} operators. We illustrate how we integrate the SCORE update (\Cref{eq:score_update}), by applying it after these operations have been computed by the base CQA model, with $\boldsymbol{\Delta}_s$ indicating similarity scores.}
    \label{fig:score_update}
\end{figure}

As described in~\Cref{sec:related_work}, several neuro-symbolic methods exist for CQA. A common approach consists of computing one \textbf{fuzzy vector} of scores $\mathbf{v}_i\in [0, 1]^{\vert\entities\vert}$ per variable in a query $\query$. Each entry in the vector indicates the likelihood of each entity in $\entities$ to be assigned to $v_i$. Some examples of methods following this approach are CQD~\citep{arakelyan2021cqd,arakelyan2023cqdA}, QTO~\citep{bai2023qto}, GNN-QE~\citep{zhu_neural-symbolic_2022}, and UltraQuery~\citep{galkin_foundation_2024}.
These methods rely on \emph{projection} and \emph{intersection} operators to model the logical query with fuzzy vectors, as illustrated in~\Cref{fig:score_update}. Once a CQA method has processed a query, the fuzzy vector for the target variable is used as the score for each entity in the KG to answer the query. We describe the fundamental operations used by CQA methods to compute fuzzy vectors for a query in~\Cref{app:neurosymbolic_methods}. Here, we focus on how to incorporate
similarity constraints into this computation.

\paragraph{Similarity-constrained Reweighting (SCORE).} The key difference between CQA and SimCQA is the inclusion of a binary predicate in the constrained answer set that restricts one of the variables $v_j$ to meet the similarity constraint $s$. In principle, this suggests modifying the fuzzy vector $\mathbf{v}_j$ computed by an underlying CQA model by increasing or decreasing the score for each entity depending on whether $s(e)$ is true or false. There are, however, two challenges: (i) the true similarity constraint $s$ is unobserved, and instead it must be approximated from the preference set $P_s$, and (ii) any adjustments made to a fuzzy vector will have an effect in the final answers to the query as the vector is propagated by the underlying CQA method throughout the rest of the query.

We address these challenges with Similarity-constrained Reweighting (SCORE). Intuitively, given a preference set $P_s$ and a fuzzy vector $\mathbf{v}_j$ computed by a base CQA model, SCORE promotes entities that resemble positively labeled, and demotes those resembling negative ones. As we illustrate in~\Cref{fig:score_update}, we apply updates to a fuzzy vector after the base CQA model has applied a projection or intersection operator. This \emph{reweighted} vector is then propagated through the rest of the query using the base CQA model.

To approximate the true similarity constraint, SCORE relies on entity embeddings obtained from the base CQA model. We employ these embeddings to compute the cosine similarity $\cosim(e_1,e_2)\in[-1, 1]$ between pairs of entities $e_1,e_2\in\entities$. In order to combine this value with fuzzy vectors whose elements lie in the interval $[0, 1]$, we compute a \emph{normalized similarity score}, defined as follows:
\begin{equation}
    \ncosim(e_1, e_2) = \tfrac{1}{2}\bigl(\cosim(e_1, e_2) + 1\bigr).
\end{equation}
This transformation captures the notion that $\ncosim(e_1, e_2)\in[0, 1]$ tends to 0 if the entities are not similar, and to 1 if they are similar.

SCORE computes updates to a fuzzy vector in \emph{logit space}. For a probability $p$, the logit is defined as $\logit(p) = \log(\tfrac{p}{1-p})$. Logits provide an additive scale for combining independent sources of evidence, which stands in contrast with multiplicative updates that can vanish quickly when values are in the interval $[0, 1]$. In preliminary experiments, a multiplicative update in probability space indeed performed worse, as scores vanished when several similarity terms were close to zero, especially when constraints were applied before further query operations.

We define $\prefpos$ as the set of entities $e$ in $P_s$ for which $s(e_i)= \text{true}$, and similarly $\prefneg$ for the case where $s(e_i)=\text{false}$. The SCORE update consists of two steps:

\begin{enumerate}
    \item For each entity $e\in\entities$, compute the mean logit similarity:
    \begin{equation}
        \Delta(\prefvar, e) = \frac{1}{\vert\prefvar\vert} \sum_{e_i\in\prefvar}\logit(\ncosim(e_i, e)),
    \end{equation}
    where $\prefvar\in\{\prefpos,\prefneg\}$.

	    \item Reweigh the score of each entity based on the mean logit similarities in logit space, and transform back to the interval $[0, 1]$ with the sigmoid function:
	    \begin{equation}
	        \mathbf{v}^{\text{new}}_j[e] = \sigma\bigl(\logit(\mathbf{v}_j[e]) + w_p \Delta(\prefpos, e) - w_n \Delta(\prefneg, e)\bigr).
	    \label{eq:score_update}
	    \end{equation}
\end{enumerate}

SCORE thus only contains two hyperparameters $w_p,w_n\in\mathbb{R}$ which can be tuned  to balance the contributions from preferred and non-preferred sets of entities, using a small validation set.

\subsection{Theoretical properties}
\label{sec:theoretical_properties}

We now highlight three important properties of SCORE. We refer to~\Cref{app:additional_props} for proofs of the propositions and details on the connection with Bayesian inference.

\begin{restatable}[Monotonicity of SCORE updates]{proposition}{Monotonicity}
For any two entities $e_1,e_2\in\entities$, suppose their preference contributions are equal:
\begin{equation}
    w_p \Delta(\prefpos, e_1) - w_n \Delta(\prefneg, e_1) =\; w_p \Delta(\prefpos, e_2) - w_n \Delta(\prefneg, e_2).
\end{equation}
Then the relative ordering of their fuzzy scores is preserved under the SCORE update:
\[
    \mathbf{v}_j[e_1] < \mathbf{v}_j[e_2]
    \;\;\Longrightarrow\;\;
    \mathbf{v}_j^{\text{new}}[e_1] < \mathbf{v}_j^{\text{new}}[e_2].
\]
\end{restatable}

The monotonicity of SCORE contributes to interpretability: whenever the score of entity is promoted or demoted, the exact source of this shift can be traced back to $\Delta(\prefpos, e)$ and $\Delta(\prefneg, e)$. This makes it transparent whether the change is due to evidence from positively or negatively labeled examples, providing a direct explanation of how similarity constraints affect the final scores.

\begin{restatable}[Linear complexity of SCORE]{proposition}{LinearComplexity}
The computational complexity of applying SCORE is $O(|\entities|)$, i.e. linear with respect to the number of entities.
\end{restatable}

\paragraph{Connection with posterior log-odds.} The SCORE update reflects the log-odds update rule in Bayesian inference. This provides a principled justification for our formulation, in which preferences act as additional likelihood terms that adjust the scores of the base CQA model in a coherent way.

\section{Experiments}
\label{sec:experiments}

\begin{figure*}[t]
    \centering
    \begin{subfigure}{\textwidth}
        \centering
        \includegraphics[width=\textwidth]{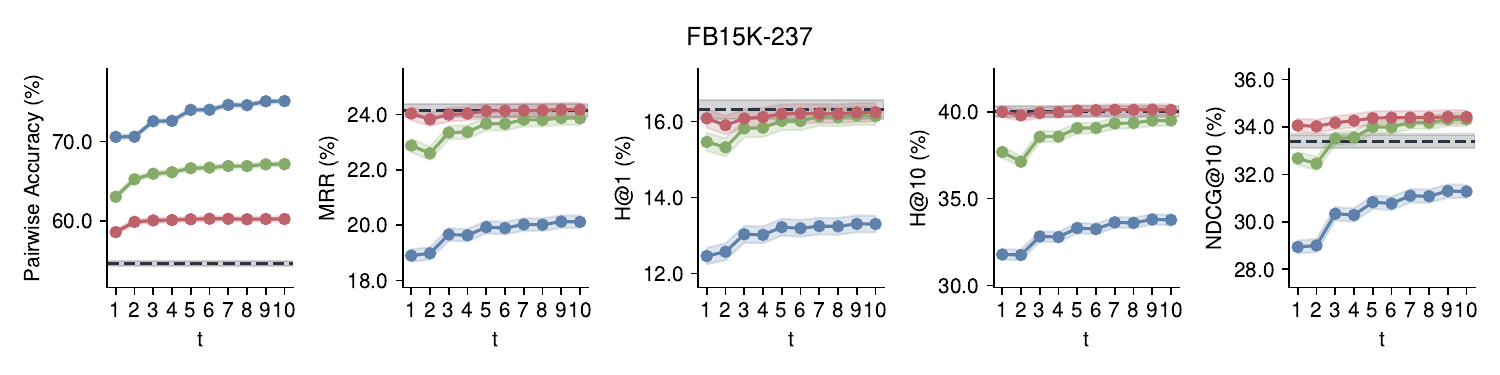}
        \includegraphics[width=\textwidth]{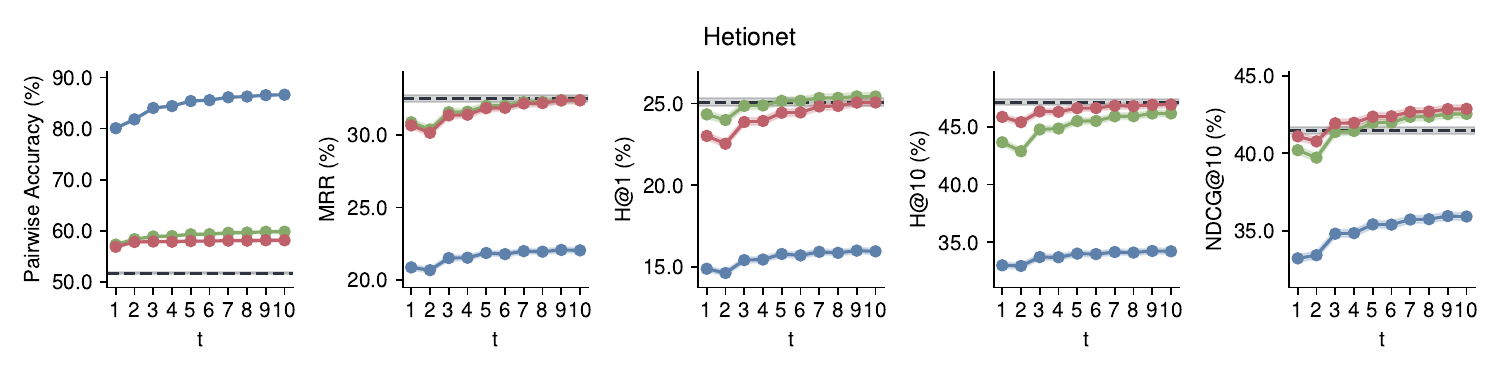}
        \includegraphics[width=\textwidth]{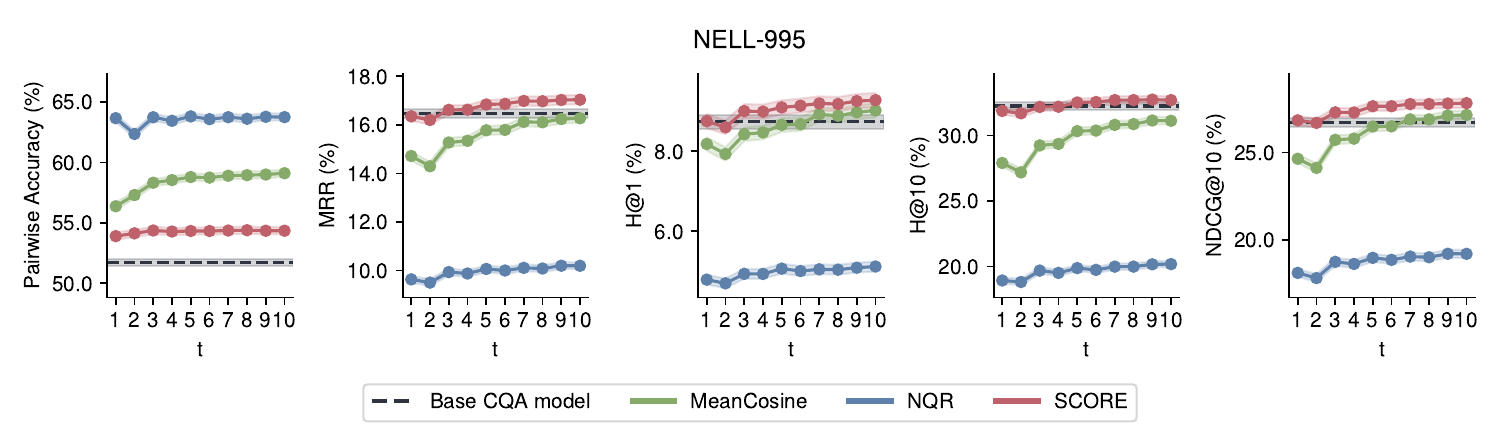}
    \end{subfigure}
    \caption{SimCQA results for similarity constraints on the \textbf{target} variable. Shaded areas indicate a 95\% confidence interval.}
    \label{fig:target_overtime}
\end{figure*}

In our experiments, we follow the ranking-based evaluation which has been employed in works on link prediction~\citep{ji_survey_2020} and complex query answering~\citep{ren2024neural}. In this setting, the set of triples $\edges$ is split into disjoint sets $\edgestrain$, $\edgesvalid$, and $\edgestest$, and the answer set $A_i(q)$ contains answers that are only reachable by traversing edges in $\edgestest$. The goal is to determine the rank assigned by a model to the entities in the answer set $A_i(q)$ when given access to the edges in $\edgestrain\cup\edgesvalid$. Similar to prior work~\citep{daza2025interactive}, in SimCQA the goal is the same, with the addition that entities in the \emph{constrained} answer set $\hat{A}_i(q,s,j)$ should be ranked higher than those not in it.

\paragraph{Datasets.} We conduct experiments on knowledge graphs of different domains and scales:
Hetionet~\citep{himmelstein_systematic_2017} is a biomedical KG covering genes, diseases, and drugs, among other biomedical entities. 
FB15k-237~\citep{bollacker_freebase_2008,toutanova_observed_2015} and NELL995~\citep{carlson_toward_2010} are encyclopedic KGs that contain general facts about people, organizations, and locations. We adopt the complex queries of~\citet{ren_beta_2020}, which cover a wide variety of 14 types of complex queries involving conjunctions, disjunctions, and negations.

In~\citet{daza2025interactive}, the authors build a benchmark of preference sets from clusters of query answers derived from textual embeddings, which in a human validation subset results in 79\% agreement. This agrees with findings that textual embedding similarity correlates with human judgments of similarity~\citep{2019reimerssbert,grand2022semantic,merk2022seeing}. We extend instances in this benchmark with constraints on arbitrary variables. We present additional details of the datasets in~\Cref{app:datasets}.

\paragraph{Baselines.} We consider three competitive methods as base CQA models: QTO~\citep{bai2023qto}, GNN-QE~\citep{zhu_neural-symbolic_2022}, and UltraQuery~\citep{galkin_foundation_2024}. Given a base CQA model, we apply SCORE to compute score updates for the SimCQA problem, in two settings where (1) similarity constraints are applied to the target variable (\textbf{target-variable} SimCQA), and (2) to any variable (\textbf{general} SimCQA).

We compare the performance of SCORE against NQR~\citep{daza2025interactive}, a neural reranker that, unlike SCORE, requires training with hundreds of thousands of instances of queries and their corresponding constrained answer sets; and MeanCosine: a simple baseline based on raw cosine similarity values, which modifies the fuzzy vector by adding/subtracting the mean cosine similarity with respect to positive/negative examples~\citep{daza2025interactive}. For general SimCQA, we adapt MeanCosine by applying its update to the intermediate fuzzy vector and mapping the result to $[0,1]$ with a sigmoid before query execution continues.
To ensure a fair comparison, we run an extensive hyperparameter grid search for SCORE, NQR, and MeanCosine separately for each dataset, selecting the best values on the validation set. More experimental details are available in~\Cref{app:exp_details}.

\paragraph{Metrics.} We evaluate each method in an \textbf{interactive setting}~\citep{daza2025interactive}, where preference sets $P_s$ of increasing size $t=1,\ldots,10$ are provided and performance is measured at each step.
We partition the answer set $A_i(q)$ into answers in the constrained answer set $\hat{A}_i(q,s,j)$, which we denote as $A_i(q)^{+}$, and those not in it, denoted as $A_i(q)^{-}$. We report the following (see~\Cref{app:metrics} for details):

\begin{itemize}
    \item \textbf{MRR} and \textbf{Hits@k (H@k)} measure whether a method preserves the ranking quality of the base CQA model, i.e., how well entities in the unconstrained answer set $A_i(q)$ are ranked. 
    \item \textbf{Pairwise Accuracy (PA)} isolates similarity-constraint satisfaction, assessing if entities in the constrained set $\hat{A}_i(q,s,j)$ are ranked above those in $A_i(q)^{-}$. 
    \item \textbf{NDCG@k} combines both aspects by assigning graded relevance (0 to non-answers, 1 to $A_i(q)^{-}$, and 2 to $A_i(q)^{+}$), rewarding a balance of global quality with constraint satisfaction.
\end{itemize}

\subsection{Results}

\begin{figure*}[t]
    \centering
    \begin{subfigure}{\textwidth}
        \centering
        \includegraphics[width=\textwidth]{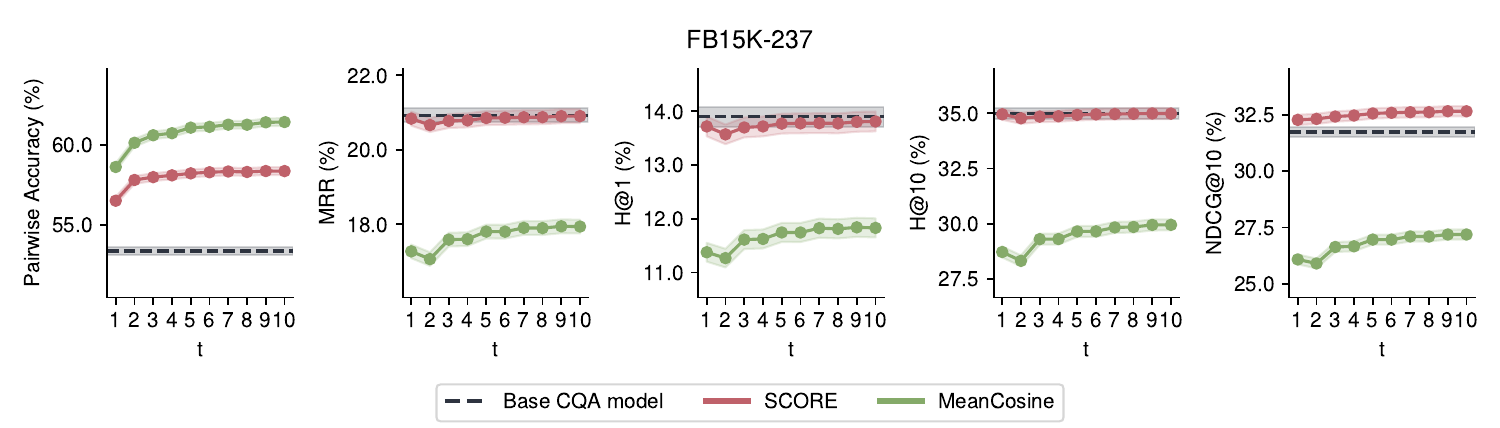}
        \includegraphics[width=\textwidth]{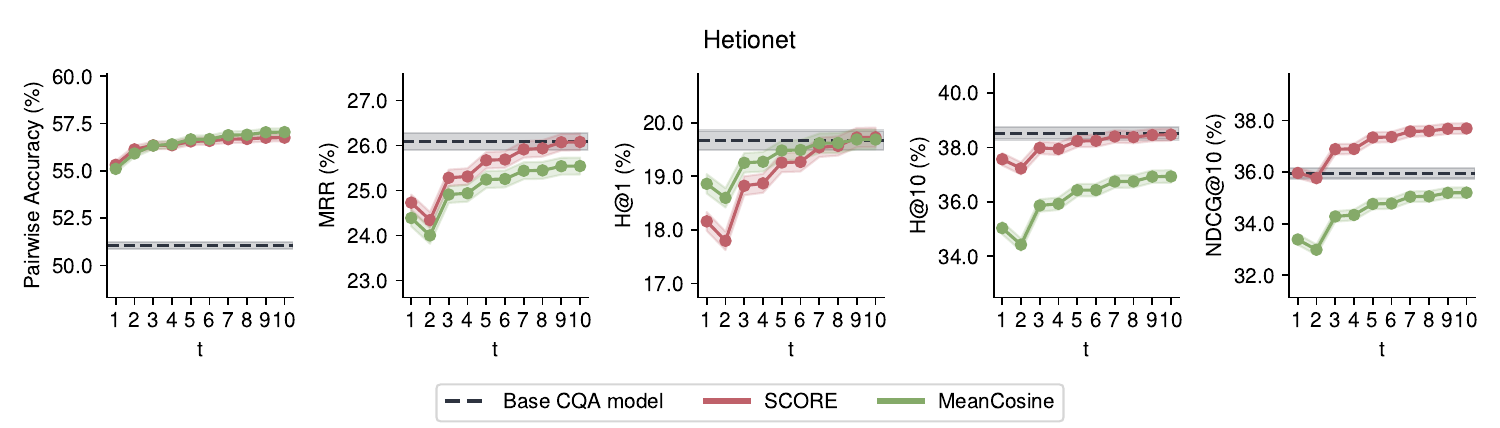}
        \includegraphics[width=\textwidth]{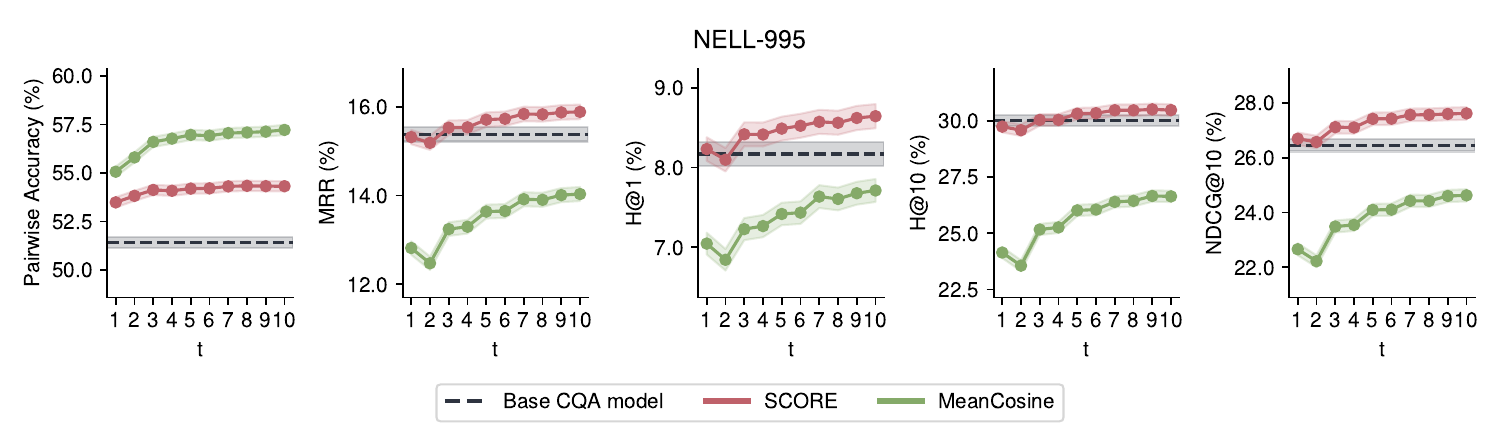}
    \end{subfigure}
    \caption{SimCQA results for similarity constraints on the \textbf{target and intermediate} variables, comparing SCORE with the adapted MeanCosine baseline. Shaded areas indicate a 95\% confidence interval.}
    \label{fig:full_overtime}
\end{figure*}

\paragraph{Target-variable SimCQA.}

We present results for target-variable SimCQA in~\Cref{fig:target_overtime}, averaged across all queries and shown with 95\% confidence intervals, using QTO~\citep{bai2023qto} as the \textbf{base CQA model}.

SCORE achieves the highest NDCG@10 across all datasets, indicating that its resulting ranking more closely resembles the optimal ranking where preferred answers are ranked above non-preferred ones, which are in turn ranked higher than entities not answering a query. Its MRR and H@10 remain close to the base CQA model, which is desirable: SCORE improves preference satisfaction without disrupting the base model's answer ranking. SCORE performs particularly well on small preference sets, indicating that it can quickly adapt to a few labeled examples.

The MeanCosine baseline shows a competitive NDCG@10, but for small preference sets, SCORE generally exhibits larger performance. This is desirable as it indicates that SCORE requires fewer entities in a preference set to capture preferences while retaining global ranking quality. While MeanCosine results in higher pairwise accuracy, we observe that it comes at the cost of degradation in ranking quality, based on the lower values of MRR and H@10. H@1 and H@3 results (the latter included in~\Cref{app:additional_results}) confirm that SCORE also preserves top-rank answer quality.

The neural NQR model achieves high pairwise accuracy, indicating strong preference satisfaction, but it does so in a more aggressive way that harms global ranking quality. While NQR is designed to balance the two goals~\citep{daza2025interactive}, these results suggest that NQR can over-optimize local similarity signals at the expense of global ranking fidelity, whereas SCORE preserves ranking structure more reliably.

\paragraph{General SimCQA.} In this setting, preferences may be applied before the target variable is reached, so score updates must remain compatible with the base CQA model as query execution continues. As shown in~\Cref{fig:full_overtime}, SCORE steadily improves NDCG@10 as more preference examples are provided, indicating that preferred answers move up in the ranking while global answer quality is preserved. The adapted MeanCosine baseline captures pairwise preferences more aggressively, but substantially degrades global ranking quality: on FB15k-237 at $t=10$, it raises pairwise accuracy from 53.3 to 61.4 but lowers MRR from 20.9 to 17.9, whereas SCORE reaches 58.4 while preserving the MRR of 20.9. This indicates that bounding unnormalized updates with a sigmoid is insufficient for reliable propagation through fuzzy operators. For reference, an oracle that places preferred answers above non-preferred ones while preserving the base ranking otherwise would reach 100\% pairwise accuracy at the base MRR. SCORE closes 10.9\% of this gap without any training.

\paragraph{Robustness across base CQA models.} We present further results on both target-variable and general SimCQA when replacing the base CQA model with GNN-QE~\citep{zhu_neural-symbolic_2022} and UltraQuery~\citep{galkin_foundation_2024} in~\Cref{app:additional_results} (\Cref{fig:target_overtime_gnnqe_ultra,fig:full_overtime_gnnqe_ultra}
). As in our previous results, SCORE improves preference satisfaction and ranking metrics with growing preference sets. The gains are larger with GNN-QE and UltraQuery than with QTO, indicating that these models benefit more from SCORE.

\paragraph{Per-query type results.} We present detailed results for each of the 14 query types in~\Cref{app:additional_results}. SCORE performs best for the large majority of query structures; the few cases where MeanCosine is slightly better do not form a consistent trend across datasets.

\paragraph{Noisy preferences and embedding quality.} In order to study the robustness of SCORE to noisy preference sets, we perform an experiment on FB15k-237 where we flip the labels of a fraction of the entities in each preference set. We observe that SCORE remains above the base CQA model in pairwise accuracy with up to 40\% flipped labels, while MRR is essentially unchanged. We provide additional details in~\Cref{app:noise}. Similarly, to study how SCORE depends on the quality of entity embeddings, we measure how well the embedding space separates preferred from non-preferred entities. We observe that this alignment is weakly but consistently associated with gains in pairwise accuracy, while it is not associated with MRR. We provide additional details in~\Cref{app:alignment}.

\paragraph{Multiple constraints.} In order to understand whether SCORE extends to multiple simultaneous constraints, we perform an experiment on FB15k-237 where similarity constraints are applied to two variables of 2p and 3p queries at the same time. We observe that SCORE improves pairwise accuracy while MRR remains close to the base CQA model. We provide additional details in~\Cref{app:multiple_constraints}.

\begin{figure*}[t]
    \centering
    \includegraphics[width=\linewidth]{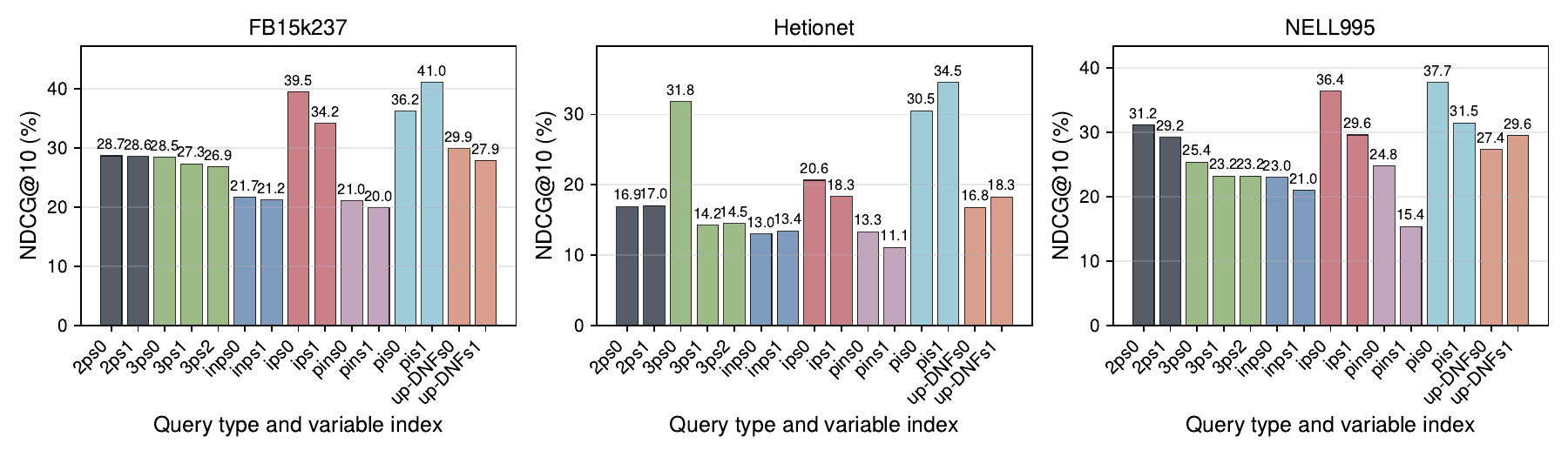}
    \caption{NDCG@10 per query type and variable position on different datasets. 
Indices indicate where the positions of variables in which similarity constraints are applied in the query.}
    \label{fig:multivariable_results}
\end{figure*}

\begin{table*}[t]
    \caption{Example 2-hop query from Hetionet, showing the top-scored entities for both the intermediate (drug) and target (side effect) variables before and after applying similarity constraints with SCORE. \promoted~indicates a promoted entity, \demoted~a demoted one.}
    \label{tab:example_rerank}
    \centering
    \small
    \begin{tabular}{m{0.22\textwidth}m{0.22\textwidth}m{0.22\textwidth}m{0.22\textwidth}}
    \toprule
    \multicolumn{4}{c}{$q(v_1,v_2) = \texttt{TreatedBy(\texttt{Migraine}}, v_1)\wedge\texttt{SideEffect}(v_1,v_2)$} \\
    \multicolumn{4}{c}{
    $P_s^{+}=\{$Diclofenac, Naproxen, Prednisone$\}$,
    $P_s^{-}=\{$Rizatriptan, Orphenadrine, Ergotamine$\}$. } \\
    \midrule
    \mc{2}{c}{\textbf{Initial Top-5}} & \mc{2}{c}{\textbf{Top-5 after SCORE}} \\
    $v_1$        & $v_2$                & $v_1$ & $v_2$ \\
    \cmidrule(lr){1-2} \cmidrule(lr){3-4}
    Sumatriptan  & Hemiplegia transient  & \promoted~Diclofenac   & \promoted~Hypersensitivity \\
    Ergotamine   & Cluster headache      & \promoted~Naproxen     & \promoted~Dermatitis \\
    Frovatriptan & Arteriospasm coronary & \promoted~Prednisone   & \promoted~Pain \\
    Antipyrine   & Temporal arteritis    & \demoted~Sumatriptan  & \promoted~Headache \\
    Naratriptan  & Hypertensive episode  & \demoted~Frovatriptan & \promoted~Asthma \\
    \midrule
                 & NDCG@10: 28.4        &              & NDCG@10: 36.5 \\
    \bottomrule
    \end{tabular}
\end{table*}

\begin{wraptable}[5]{r}{0.4\textwidth}
\vspace{-20pt}
\centering
\small
\caption{Average runtime per query (ms) with and without SCORE.}
\label{tab:runtime}
\resizebox{0.38\textwidth}{!}{
\begin{tabular}{lrrr}
\toprule
Dataset & Base & +SCORE & $\Delta$ \\
\midrule
FB15k237 & 2.96 & 4.20 & +1.24 \\
Hetionet & 25.65 & 28.68 & +3.03 \\
NELL-995 & 75.51 & 83.96 & +8.45 \\
\bottomrule
\end{tabular}
}
\end{wraptable}
\paragraph{Runtime overhead.} \Cref{tab:runtime} reports average query execution times (in ms) with and without SCORE. The additional cost is modest (from 1.24 to 8.45 ms per query) and does not grow superlinearly with dataset size, confirming that SCORE preserves the efficiency of existing CQA methods.

\paragraph{Influence of variable position.}
\Cref{fig:multivariable_results} reports results by types of queries containing more than one variable, and by the variable position on which similarity constraints are applied. 
We specify an index that indicates the topological order of variables in the query (see~\Cref{fig:queries} in~\Cref{app:datasets}). 
For example, in a 2-hop query, \texttt{2ps0} applies similarity constraints to the intermediate variable, while \texttt{2ps1} applies them to the target variable. 
In most cases, constraints applied \emph{earlier} in the query (i.e., on intermediate variables) yield slightly higher performance than when applied to later variables. 
This suggests that preference information is more effectively propagated when introduced at earlier stages of the query, reinforcing the value of supporting constraints on arbitrary variables beyond the target.

\paragraph{Qualitative example.} \Cref{tab:example_rerank} illustrates how SCORE applies similarity constraints to an intermediate variable. The base CQA model (QTO in this example) retrieves drugs from triptan and ergot classes, together with unrelated side effects, reaching an NDCG@10 of 28.4. With preferences favoring anti-inflammatory drugs and steroids, SCORE shifts the ranking to those compounds and their associated side effects, improving NDCG@10 to 36.5.

\paragraph{Limitations and future work.} In this work, similarity constraints are specified through example-based preference sets, and SCORE approximates them with cosine similarities between entity embeddings. Its effectiveness thus depends on how well the embedding space reflects the intended preference and on the quality of the examples: SCORE tolerates moderate label noise, but predominantly misleading feedback reduces preference satisfaction. A promising direction for future work is to support preferences expressed directly in natural language, including hybrid pipelines where LLM- or tool-guided traversal proposes candidates before SCORE reweights full entity score vectors. In addition, it remains interesting to investigate how to incorporate similarity constraints into methods that embed an entire query into a vector, also known as Skolemization methods~\citep{lu2025exploring}.

\section{Conclusion}

We introduce and formalize the general problem of Similarity-constrained Complex Query Answering (SimCQA), extending logical queries with similarity constraints on arbitrary variables. To address this problem, we propose Similarity-Constrained Reweighting (SCORE), a computationally efficient and interpretable reweighting layer for base CQA models that propagate entity score vectors. Experiments across multiple knowledge graphs and base CQA models show that SCORE progressively incorporates similarity constraints as preference sets grow, improving preference satisfaction while preserving or improving query answering performance. We further find that constraints applied earlier in the query often yield stronger gains by allowing preference information to propagate more effectively. These results highlight SCORE as a practical and general approach for specifying constraints beyond first-order logic in CQA.

\newpage

\section*{Acknowledgments}

Daniel Daza was funded by a gift from Accenture LLP.
The work by Michael Cochez has received funding from the European Union under the CEREBRA-AI project (Grant Agreement No. 101299046). Views and opinions expressed are, however, those of the author(s) only and do not necessarily reflect those of the European Union or the European Health and Digital Executive Agency (HaDEA). Neither the European Union nor HaDEA can be held responsible for them.
Work on this publication is in part based upon work from COST Action CA23147 GOBLIN
- Global Network on Large-Scale, Cross-domain and Multilingual Open Knowledge Graphs, and COST Action
CA24121 - Knowledge Graphs in the Era of Large Language Models (KGELL), both supported by COST
(European Cooperation in Science and Technology, https://www.cost.eu)

\bibliographystyle{plainnat}
\bibliography{references}


\appendix

\section{Appendix}

\subsection{Neuro-symbolic methods for CQA}
\label{app:neurosymbolic_methods}

This section provides an overview of the logical operators in the logical queries we consider in this work, as well as the relaxed operators used to model logical operators in existing methods for neuro-symbolic CQA that can be used as base methods for SimCQA.    

\paragraph{Logical operators.} 
We consider logical queries over a KG $\graph = (\entities,\relations,\edges)$ specified by a formula $\query$ in first-order logic, written in disjunctive normal form (DNF), i.e., as a disjunction ($\vee$) of conjunctions ($\wedge$) of \emph{atoms}:
\begin{equation}
    \query = (c^1_{1}\land\dots\land c^1_{m_1}) \lor \dots \lor (c^n_{1}\land\dots\land c^n_{m_n}),
\end{equation}
where each $c^i_j$ denotes a predicate applied to two variables, or a variable and a known entity in $\entities$, while optionally including negations:
\begin{align}
c^i_{j} = 
\begin{cases}
    r(e, v)\text{ or }r(v', v) \\
    \lnot r(e, v)\text{ or }\lnot r(v', v)
\end{cases}
\end{align}
with $e\in\entities, v,v'\in\lbrace v_1, \ldots, v_N\rbrace, \text{ and } r\in\relations$.

\paragraph{T-norms.} 
Neuro-symbolic methods for CQA incorporate logical operators in their computations by relaxing them into continuous functions known as \textbf{t-norms} and \textbf{t-conorms} from fuzzy logics~\citep{zadeh1965fuzzy}, where values in $[0,1]$ model degrees of truth.
In this setting, conjunction ($\land$) is generalized by a \emph{t-norm} $\top:[0,1]^2\to[0,1]$, disjunction ($\lor$) by a \emph{t-conorm} $\bot:[0,1]^2\to[0,1]$, and negation ($\lnot$) by the standard fuzzy complement $1-x$. Using these relaxations, a query $\query$ in disjunctive normal form can be expressed as
\begin{equation}
    \query = (c^1_{1}\top\dots\top c^1_{m_1}) \bot \dots \bot (c^n_{1}\top\dots\top c^n_{m_n}),
\end{equation}
where each atom $c^i_j$ is mapped to a fuzzy truth value in $[0,1]$.

An example is the \emph{product} t-norm and its dual t-conorm, defined as
\begin{align}
    \top(x,y) &= x \cdot y, \\
    \bot(x,y) &= x + y - x \cdot y,
\end{align}
which recover classical Boolean logic when $x,y \in \{0,1\}$. Several other t-norms exist, for an in-depth description see~\citet{hajek1998metamathematics}.

\paragraph{Fuzzy vectors for CQA.} Several neuro-symbolic methods assign a vector $\mathbf{v}_i\in[0,1]^{\vert\entities\vert}$ to each variable $v_i$ in a query, which represents the likelihood that the variable is assigned each entity in $\entities$~\citep{arakelyan2021cqd,arakelyan2023cqdA,bai2023qto,zhu_neural-symbolic_2022,cucumides_unravl_2024,yin_rethinking_2024}. Similarly, constants in a query can be represented as one hot vectors $\mathbf{c}\in\lbrace0, 1\rbrace^{\vert\entities\vert}$, such that $\mathbf{c}[e] = 1$ for a constant entity $e$, and all other entries are zero. This vector can then be transformed with t-norms, t-conorms, and negations depending on the query being answered.

\paragraph{Projections and negations.} For a query formula $q(v_1,v_2) = r(v_1, v_2)$ with $r\in\relations$, a fuzzy vector $\mathbf{v}_1$ is projected via a function $f_r(\mathbf{v}_1) = \mathbf{v}_2\in[0, 1]^{\vert\entities\vert}$ which yields the fuzzy vector for $v_2$. Often the function $f_r$ is a parameterized neural network optimized for 1-hop link prediction~\citep{arakelyan2021cqd,arakelyan2023cqdA,bai2023qto} or CQA~\citep{zhu_neural-symbolic_2022,galkin_foundation_2024}. For a negation $q(v_1,v_2) = \lnot r(v_1,v_2)$, the fuzzy vector is computed as $\mathbf{1} - f_r(\mathbf{v}_1)$, where $\mathbf{1}$ is an all-ones vector of length $\vert\entities\vert$.

\paragraph{Conjunctions and disjunctions.} For a conjunction $q(v_1, v_2, v_3) = r_1(v_1, v_2)\wedge r_2(v_3, v_2)$ with $r_1,r_2\in\relations$ the two predicates are first projected as $f_{r1}(\mathbf{v}_1)$ and $f_{r2}(\mathbf{v}_3)$, and the fuzzy vector of $\mathbf{v}_2$ is computed via the t-norm: $\mathbf{v}_2 = f_{r1}(\mathbf{v}_1) \top f_{r1}(\mathbf{v}_3)$. For a disjunction $r_1(v_1, v_2)\wedge r_2(v_3, v_2)$ the result is computed via the t-conorm, $\mathbf{v}_2 = f_{r1}(\mathbf{v}_1) \bot f_{r1}(\mathbf{v}_3)$.

\subsection{Proofs of properties of SCORE}
\label{app:additional_props}

\Monotonicity*

\begin{proof}
By definition of the update in \Cref{eq:score_update},
\[
\mathbf{v}^{\text{new}}_j[e] = \sigma\!\left(\logit(\mathbf{v}_j[e]) + \delta(e)\right),
\]
where $\delta(e) = w_p \Delta(\prefpos, e) - w_n \Delta(\prefneg, e)$.  
If $\delta(e_1) = \delta(e_2)$, then the same shift is applied to both $\logit(\mathbf{v}_j[e_1])$ and $\logit(\mathbf{v}_j[e_2])$. Since $\sigma(\cdot)$ is strictly monotone, the ordering of the original logits is preserved, and so is the ordering of the values in the fuzzy vector. Hence SCORE is monotonic with respect to the scores of the underlying model: it preserves the base model’s ranking whenever preference contributions are equal.
\end{proof}

The monotonicity of SCORE contributes to interpretability: whenever the score of entity is promoted or demoted, the exact source of this shift can be traced back to $\Delta(\prefpos, e)$ and $\Delta(\prefneg, e)$. This makes it transparent whether the change is due to evidence from positively or negatively labeled examples, providing a direct explanation of how similarity constraints affect the final scores.

\LinearComplexity*

\begin{proof}
Let $|\entities|$ be the number of entities and $k$ the number of examples in the preference set.  
For each candidate entity $e\in\entities$, SCORE computes normalized similarities $\ncosim(e, e_i)$ with all $k$ labeled examples, which costs $O(kd)$ if embeddings are $d$-dimensional. Aggregating these values and computing the update in \Cref{eq:score_update} also requires $O(k)$. Thus the cost per entity is $O(kd)$, and the total cost is $O(|\entities|kd)$. Since $k$ is small (on the order of tens of examples) and $d$ is fixed by the underlying embedding model, both factors are constant in practice. Therefore the overall complexity scales linearly with $|\entities|$.
\end{proof}

\paragraph{Connection with posterior log-odds.} Assuming a probabilistic interpretation of scores, we reinterpret the similarity constraint $s(e)$ (previously a binary predicate) as the probability that an entity $e\in\entities$ is to satisfy the constraint. For brevity, let $\mathbf{v} = \mathbf{v}_j[e]$ denote the base CQA score for entity $e$. The SCORE update can then be written as
\begin{equation}
    \mathbf{v}^{\text{new}} =  \sigma\!\left(\logit(\mathbf{v}) + \logit(s(e)) \right).
\end{equation}

Taking logits gives
\begin{equation}
    \logit(\mathbf{v}^{\text{new}}) =  \logit(\mathbf{v}) + \logit(s(e)),
\end{equation}
and by the definition of the logit,
\begin{equation}
    \log\frac{\mathbf{v}^{\text{new}}}{1-\mathbf{v}^{\text{new}}}
    = \log\frac{\mathbf{v}}{1-\mathbf{v}} + \log\frac{s(e)}{1-s(e)}.
\end{equation}

This is equivalent to the additive update of log-odds in Bayesian inference: the posterior log-odds equal the prior log-odds plus the log-likelihood ratio of the evidence. More concretely, let $x$ be the random variable indicating whether $e$ is an answer. If we view $\mathbf{v}=P(x=1)$ as the prior from the CQA model and $s(e)$ as a data-dependent score that encodes the strength of evidence for $x=1$ versus $x=0$, then
\begin{equation}
    \log \frac{P(x=1 \mid \text{data})}{P(x=0 \mid \text{data})}
    = \log \frac{P(x=1)}{P(x=0)} +
      \log \frac{P(\text{data}\mid x=1)}{P(\text{data}\mid x=0)}.
\end{equation}

Thus, the SCORE update mirrors the Bayesian log-posterior update when scores are interpreted probabilistically. While the base CQA model and similarity function may not produce calibrated probabilities, operating in logit space ensures that preference evidence is combined additively and propagated consistently with the logical operators of the query.

\subsection{Datasets}
\label{app:datasets}

\begin{figure*}[t]
    \centering
    \includegraphics[width=\linewidth]{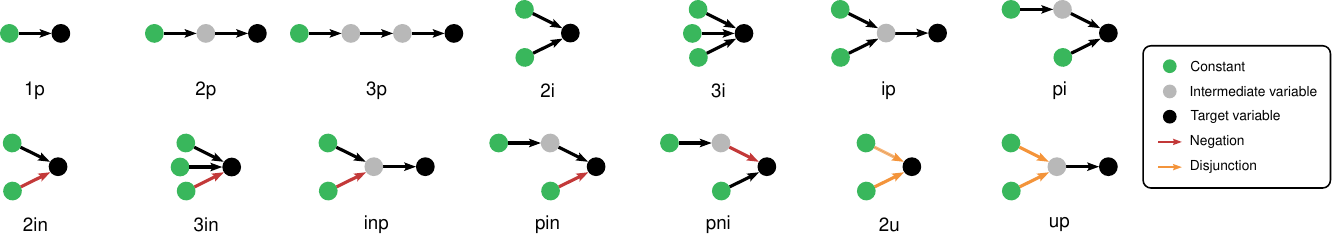}
    \caption{Query graphs of different types covered in the datasets used in our experiments, along with their designated name.}
    \label{fig:queries}
\end{figure*}

\begin{table*}[t]
    \caption{Statistics of the datasets used in our experiments.}
    \label{tab:statistics}
    \centering
    \resizebox{\textwidth}{!}{
    \begin{tabular}{cccccccccc}
    \toprule
                 &   \mc{3}{c}{Knowledge Graph}          & \mc{3}{c}{Queries} & \mc{3}{c}{Preference Sets}      \\
                   \cmidrule(lr){2-4}          \cmidrule(lr){5-7}    \cmidrule(lr){8-10} 
         Dataset & Entities & Relations &   Edges   & Train  & Validation &  Test  &  Train &  Validation &  Test   \\
    \midrule
         FB15k237 &  14,505 &       237 &   310,079 & 157,479 & 20,961 & 21,352 & 580,623 & 68,834 & 70,365 \\
         Hetionet &  45,158 &        24 & 2,250,198 & 91,820 & 29,475 & 29,488 & 325,443 & 99,785  & 99,284 \\
         NELL995  &  63,361 &       200 &   142,804 & 75,446 & 17,289 & 17,435 & 245,511 & 52,446 & 52,589 \\
    \bottomrule
    \end{tabular}
    }
\end{table*}

In this section we provide additional details on how we construct a generalized benchmark for SimCQA with similarity constraints on arbitrary variables. 

\paragraph{Base resources.}
We start from standard benchmarks for complex query answering~\citep{ren_beta_2020}, which cover 14 query types involving conjunction, disjunction, and negation (illustrated in~\Cref{fig:queries}). For each knowledge graph (Hetionet, FB15k-237, and NELL995), we generate training, validation, and test queries following the same procedure as in prior work. Each query is associated with a designated \emph{target variable} $v_i$ whose answers form the set $A_i(q)$.

\paragraph{Preference sets and constrained answers.}
Following~\citet{daza2025interactive}, we rely on clusterings of textual embeddings as a scalable proxy for unobserved similarity constraints. In their human validation subset on FB15k-237, these embedding-derived clusters agree with human judgments in 79\% of cases, supporting their use as preference sets for large-scale evaluation. For each query, we (i) identify a variable $v_j$ (which may be the target or any intermediate variable), (ii) collect the entities that appear as valid bindings for $v_j$ and compute an embedding of their textual descriptions, and (iii) apply hierarchical agglomerative clustering to obtain two partitions of this set. These partitions define preference sets of the form $P_s = \{(e, \texttt{true}) : e \in C^+\} \cup \{(e, \texttt{false}) : e \in C^-\}$, where $C^+$ and $C^-$ are clusters. The induced constrained answer set $\hat{A}_i(q,s,j)$ is then computed by propagating the partition of $v_j$ through the query, yielding a corresponding partition of the target answers. Textual information is used only to construct preference sets and constrained answer sets, and the evaluated SimCQA methods only access graph topology.

\paragraph{Text-based entity embeddings.} Embeddings for entities used during the clustering stage are computed using their textual description. We use pretrained text embedding models available through Hugging Face\footnote{\url{https://huggingface.co/}} to derive vector representations of entities from their textual descriptions. For FB15k237, we adopt the \texttt{NovaSearch/stella\_en\_400M\_v5} model\footnote{\url{https://huggingface.co/NovaSearch/stella_en_400M_v5}}, which produces 1{,}024-dimensional embeddings. For Hetionet, we rely on \texttt{Alibaba-NLP/gte-Qwen2-7B-instruct}\footnote{\url{https://huggingface.co/Alibaba-NLP/gte-Qwen2-7B-instruct}}. At the time of writing, this model holds the top position on the Massive Text Embedding Benchmark\footnote{\url{https://huggingface.co/spaces/mteb/leaderboard}} for medical-domain tasks and yields embeddings of dimension 3{,}584.

\paragraph{Generalization protocol.}
A key difference with~\citet{daza2025interactive} is how we split answers across $\edgestrain$, $\edgesvalid$, and $\edgestest$. In their original setup, preference examples could overlap with answers reachable in $\edgestest$, meaning that at test time the model effectively observes part of the ground truth answers (a few-shot learning scenario). To ensure stricter generalization, we impose the following rules:
\begin{itemize}
    \item The ground truth constrained answer set $\hat{A}_i(q,s,j)$ is defined only with respect to $\edgestest$. This ensures that evaluation reflects the model’s ability to generalize to unseen facts.
    \item Preference sets $P_s$ are restricted to contain only entities reachable via $\edgestrain \cup \edgesvalid$. This prevents test leakage and forces the model to extrapolate similarity constraints to novel cases at test time.
\end{itemize}

\paragraph{Constraints on arbitrary variables.}
Whereas~\citet{daza2025interactive} considered constraints only on the target variable, we generalize the construction to allow constraints on \emph{any} variable in the query. This is achieved by clustering the bindings of each free variable $v_j$ in the query (not just the target), and retaining those partitions that induce non-trivial and non-overlapping partitions of the target answer set. This extension produces preference/evidence pairs where feedback is given on intermediate variables, leading to richer scenarios for similarity-constrained reasoning.

\paragraph{Filtering and quality control.}
To ensure meaningful preference sets, we apply the following filters:  
(i) only variables with at least 10 and at most 100 distinct bindings are considered for clustering;  
(ii) both positive and negative clusters must induce at least 5 valid answers on the training edges;  
(iii) induced answer sets must be non-empty and disjoint.  
These criteria discard degenerate cases and guarantee that similarity constraints provide useful supervisory signals.

\paragraph{Output.}
Each dataset instance consists of:
\begin{enumerate}
    \item a query $q$ and target variable $v_i$,
    \item the variable $v_j$ on which preferences are applied,
    \item a preference set $P_s$ built from clusters of bindings of $v_j$,
    \item the ground truth constrained answer set for the target variable $v_i$: $\hat{A}_i(q,s,j)$ derived from $\edgestest$.
\end{enumerate}
We repeat this process across all query types and variables, producing a diverse benchmark that covers both biomedical and encyclopedic domains. Dataset statistics are reported in~\Cref{tab:statistics} and~\Cref{tab:query_stats}.

\begin{table*}[t]
\caption{Statistics of the queries and preference sets in the datasets used in our experiments.}
    \label{tab:query_stats}
    \centering
    \resizebox{\textwidth}{!}{
    \begin{tabular}{cccccccccccccccc}
    \toprule
   Structure &       1p &       2p &       3p &       2i &       3i &       ip &       pi &      2in &      3in &      inp &      pin &      pni &       2u &       up &    Total \\
   \midrule
   & \mc{15}{c}{\textbf{FB15k237}} \\
   \cmidrule(lr){2-16}
   & \mc{15}{c}{Training} \\
   \midrule
     Queries &    7,596 &   34,720 &   56,495 &   16,205 &    8,870 &        0 &        0 &    5,300 &    3,821 &    9,384 &    8,891 &    6,197 &        0 &        0 &  157,479 \\
  Pref. Sets &   19,075 &  121,420 &  240,770 &   46,302 &   26,265 &        0 &        0 &   16,536 &   11,530 &   41,128 &   38,217 &   19,380 &        0 &        0 &  580,623 \\
  \midrule
  & \mc{15}{c}{Validation} \\
  \midrule
     Queries &    1,631 &    1,746 &    2,255 &    1,341 &      753 &    1,249 &    1,501 &    1,210 &    1,560 &    2,184 &    1,825 &      782 &    1,087 &    1,837 &   20,961 \\
  Pref. Sets &    4,149 &    5,967 &    8,876 &    3,972 &    2,275 &    4,099 &    5,239 &    3,516 &    4,549 &    8,140 &    6,256 &    2,182 &    3,251 &    6,363 &   68,834 \\
  \midrule
  & \mc{15}{c}{Test} \\
  \midrule
     Queries &    1,934 &    1,776 &    2,267 &    1,347 &      814 &    1,337 &    1,399 &    1,191 &    1,553 &    2,133 &    1,844 &      814 &    1,080 &    1,863 &   21,352 \\
  Pref. Sets &    4,898 &    6,157 &    9,138 &    3,875 &    2,398 &    4,358 &    4,834 &    3,456 &    4,581 &    8,017 &    6,516 &    2,356 &    3,206 &    6,575 &   70,365 \\
  \midrule
  & \mc{15}{c}{\textbf{Hetionet}} \\
   \cmidrule(lr){2-16}
   & \mc{15}{c}{Training} \\
   \midrule
     Queries &   19,595 &   20,075 &    2,000 &   20,075 &   20,075 &        0 &        0 &    2,000 &    2,000 &    2,000 &    2,000 &    2,000 &        0 &        0 &   91,820 \\
  Pref. Sets &   66,076 &   79,488 &    9,127 &   68,553 &   66,012 &        0 &        0 &    7,281 &    6,768 &    7,503 &    7,586 &    7,049 &        0 &        0 &  325,443 \\
  \midrule
  & \mc{15}{c}{Validation} \\
  \midrule
     Queries &    9,975 &    1,500 &    1,500 &    1,500 &    1,500 &    1,500 &    1,500 &    1,500 &    1,500 &    1,500 &    1,500 &    1,500 &    1,500 &    1,500 &   29,475 \\
  Pref. Sets &   32,653 &    5,185 &    5,805 &    5,024 &    5,017 &    5,040 &    5,801 &    5,158 &    5,068 &    5,223 &    4,970 &    4,829 &    4,877 &    5,135 &   99,785 \\
  \midrule
  & \mc{15}{c}{Test} \\
  \midrule
     Queries &    9,988 &    1,500 &    1,500 &    1,500 &    1,500 &    1,500 &    1,500 &    1,500 &    1,500 &    1,500 &    1,500 &    1,500 &    1,500 &    1,500 &   29,488 \\
  Pref. Sets &   32,865 &    5,049 &    5,796 &    5,056 &    4,980 &    4,843 &    5,783 &    5,092 &    5,121 &    5,218 &    4,789 &    4,630 &    4,820 &    5,242 &   99,284 \\
  \midrule
  & \mc{15}{c}{\textbf{NELL995}} \\
   \cmidrule(lr){2-16}
   & \mc{15}{c}{Training} \\
   \midrule
     Queries &    1,871 &   17,649 &   24,859 &    4,547 &    3,034 &        0 &        0 &    3,848 &    2,459 &    5,824 &    6,016 &    5,339 &        0 &        0 &   75,446 \\
  Pref. Sets &    4,677 &   50,328 &   92,914 &   11,221 &    7,530 &        0 &        0 &   11,507 &    7,151 &   22,040 &   22,145 &   15,998 &        0 &        0 &  245,511 \\
  \midrule
  & \mc{15}{c}{Validation} \\
  \midrule
     Queries &      942 &    1,281 &    1,532 &      824 &      468 &      836 &    1,042 &    1,381 &    1,383 &    2,126 &    1,859 &    1,058 &    1,152 &    1,405 &   17,289 \\
  Pref. Sets &    2,254 &    4,043 &    5,159 &    2,292 &    1,291 &    2,410 &    2,855 &    3,945 &    4,214 &    7,229 &    6,038 &    3,046 &    3,270 &    4,400 &   52,446 \\
  \midrule
  & \mc{15}{c}{Test} \\
  \midrule
     Queries &    1,004 &    1,253 &    1,607 &      993 &      635 &      772 &    1,039 &    1,344 &    1,510 &    2,106 &    1,770 &      938 &    1,106 &    1,358 &   17,435 \\
  Pref. Sets &    2,437 &    3,917 &    5,514 &    2,659 &    1,789 &    2,077 &    2,933 &    3,860 &    4,479 &    7,182 &    5,743 &    2,618 &    3,076 &    4,305 &   52,589 \\
  \bottomrule
    \end{tabular}
    }
\end{table*}

\subsection{Metrics}
\label{app:metrics}

\paragraph{Global ranking quality.} For each entity in the unconstrained answer set $A_i(q)$, we determine its rank $r$, and compute the Mean Reciprocal Rank (MRR) and Hits@k (H@k), computed per answer as $\text{MRR}=\tfrac{1}{r}$ and $\text{H@k} = \mathbbm{1}[r \leq k]$ (where $\mathbbm{1}$ is an indicator function) and averaged over all queries.

\paragraph{Similarity constraint satisfaction.} Here we measure whether methods rank preferred answers higher than non-preferred ones. We partition the answer set $A_i(q)$ into answers in the constrained answer set $\hat{A}_i(q,s,j)$, which we denote as $A_i(q)^{+}$, and those not in it, denoted as $A_i(q)^{-}$. We then compute Pairwise Accuracy:
\begin{equation}
\label{eq:pairwise-accuracy}
    \operatorname{PA} = \sum_{e^+\in A_i(q)^{+}}\sum_{e^-\in A_i(q)^{-}} \mathbbm{1}\left[ r(e^+) < r(e^-)\right],
\end{equation}
where $r(e)$ indicates the ranking of entity $e$.

We also compute the Normalized Discounted Cumulative Gain at $k$ (NDCG@k), assigning 0 relevance to non-answers, 1 to answers in $A_i(q)^{-}$, and 2 to answers in $A_i(q)^{+}$. Concretely, the discounted cumulative gain is
\begin{equation}
    \mathrm{DCG@}k = \sum_{e\in A_i(q)} \frac{2^{\mathrm{rel}(e)} - 1}{\log_2(j+1)},
\end{equation}
where
\begin{equation}
\label{eq:relevance}
\mathrm{rel}(e) =
\begin{cases}
    0 & \text{if}\ e \notin A_i(q) \\
    1 & \text{if}\ \in A_i(q) \setminus \hat{A}_i(q,s,j) \\
    2 & \text{if}\ \in \hat{A}_i(q,s,j)
\end{cases}
\end{equation}
and the normalized score is
\begin{equation}
    \mathrm{NDCG@}k = \frac{\mathrm{DCG@}k}{\mathrm{IDCG@}k},
\end{equation}
where $\mathrm{IDCG@}k$ is the maximum achievable DCG under the given relevance scores. Since NDCG@k is normalized by the ideal ranking, a value of 1 indicates that all preferred answers appear above non-preferred answers, which in turn appear above non-answers. We report the average of PA and NDCG@k over all queries.

\subsection{Experimental details}
\label{app:exp_details}

\paragraph{Confidence intervals.} In all plots with shaded regions, we report 95\% confidence intervals for each preference-set size $t=1,\ldots,10$. Concretely, we first compute the mean and standard error of the mean (SEM) across queries, and then multiply the latter by the two-sided Student-$t$ critical value with $n-1$ degrees of freedom, where $n$ is the number of evaluated queries. This corresponds to $\mathrm{CI}_{95}=\mathrm{SEM}\cdot t_{0.975,n-1}$ after scaling metrics to percentages.

\paragraph{Compute resources.} All experiments were run on a cluster using a single NVIDIA A100 GPU with 40GB of memory, 120GB RAM, and an Intel Xeon CPU.

\paragraph{Licenses of existing assets.} FB15k-237 is distributed under the CC-BY-4.0 license, and Hetionet is distributed under CC0. NELL995 is freely available, but to the best of our knowledge there is no explicit license information for it.

\paragraph{SCORE.} The only two parameters that need to be tuned in SCORE are the weights $w_p$ and $w_n$ in~\Cref{eq:score_update} that control the strength of shifts in logit space to the base score due to similarities with the preference set. We run a grid search with values in $\{0.25, 0.5, 1.0\}$, for a total of 9 possible combinations of $w_p$ and $w_n$.

\paragraph{NQR.} We perform a grid search with values of the learning rate in $\lbrace1\times 10^{-5}, 1\times 10^{-4}\rbrace$, the margin in the preference loss in $\lbrace0.05, 0.1, 0.25\rbrace$, and the KL divergence weight in $\lbrace0.1, 1.0, 10\rbrace$, resulting in a total of 18 possible combinations of hyperparameters. As done by~\citet{daza2025interactive}, we train NQR on queries of type 1p, as training on more query types is more expensive but brings little benefits.

\paragraph{MeanCosine.} The update rule in the MeanCosine baseline is
\begin{equation}
\mathbf{v}_i^{\text{new}}[e]
= \alpha\,\mathbf{v}_i[e]
  + (1-\alpha)\Bigg(
    \frac{1+\beta}{2\lvert \prefpos\rvert}
    \sum_{e_i\in\prefpos}\cosim(e_i,e)
    -\frac{1-\beta}{2\lvert \prefneg\rvert}
    \sum_{e_i\in\prefneg}\cosim(e_i,e)
  \Bigg).
\end{equation}

Intuitively, MeanCosine computes a convex combination of the original score, and a score due to the mean of raw cosine similarity values, which is balanced with the hyperparameter $\alpha\in[0, 1]$. Similarities from $P_s^+$ are added, and those from $P_s^-$ are subtracted, and the balance of these two is determined by the value of $\beta\in[-1,1]$. We run a grid search for the values of $\alpha$ in $\{0.25, 0.5, 0.75\}$, and $\beta\in\{-0.5, 0, 0.5\}$ for a total of 9 possible configurations. In general SimCQA, the result of this update at an intermediate variable is no longer guaranteed to lie in $[0,1]$, so we pass it through a sigmoid before it is consumed by subsequent fuzzy logical operators. We select its hyperparameters with the same grid search.

\subsection{Additional results}
\label{app:additional_results}

We present more detailed results on SimCQA performance for each of the 14 types of complex queries we consider in our work (shown in~\Cref{fig:queries}).

Following our main experiments, we apply a similarity constraint to a query by using preference sets of size $t = 1,\ldots, 10$. For each value of $t$ we then compute NDCG@10 and report the average over the 10 sizes. We present results for target-variable SimCQA in~\Cref{tab:detailed_target_results} and for general SimCQA in~\Cref{tab:detailed_full_results}.

For general SimCQA, we suffix the query type with s$\ell$, where $\ell$ is the zero-based index of the variable on which the similarity constraint is applied. Variables are ordered in a topological order from the leaf nodes to the root of the query graph (see \Cref{fig:queries}). For example, for a 2p chain, the intermediate variable corresponds to index 0 (2ps0), and the terminal variable corresponds to index 1 (2ps1).

\begin{figure*}[t]
    \centering
    \begin{subfigure}{\textwidth}
        \centering
        \includegraphics[width=\textwidth]{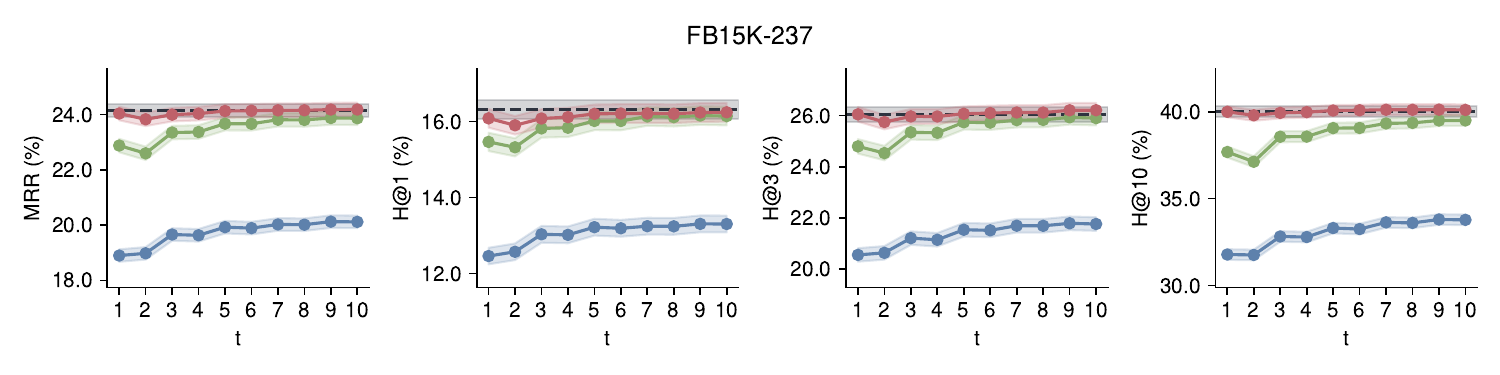}
        \includegraphics[width=\textwidth]{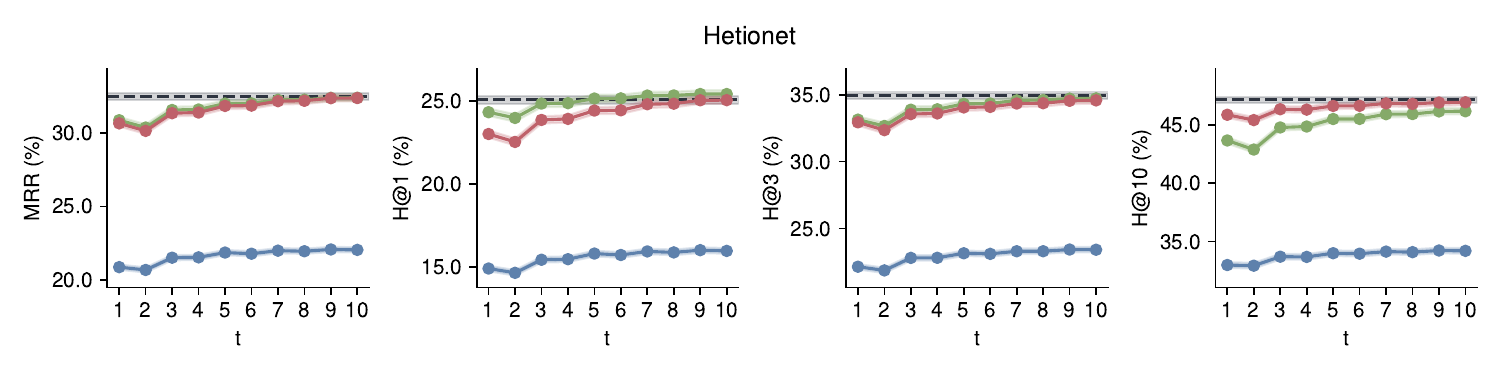}
        \includegraphics[width=\textwidth]{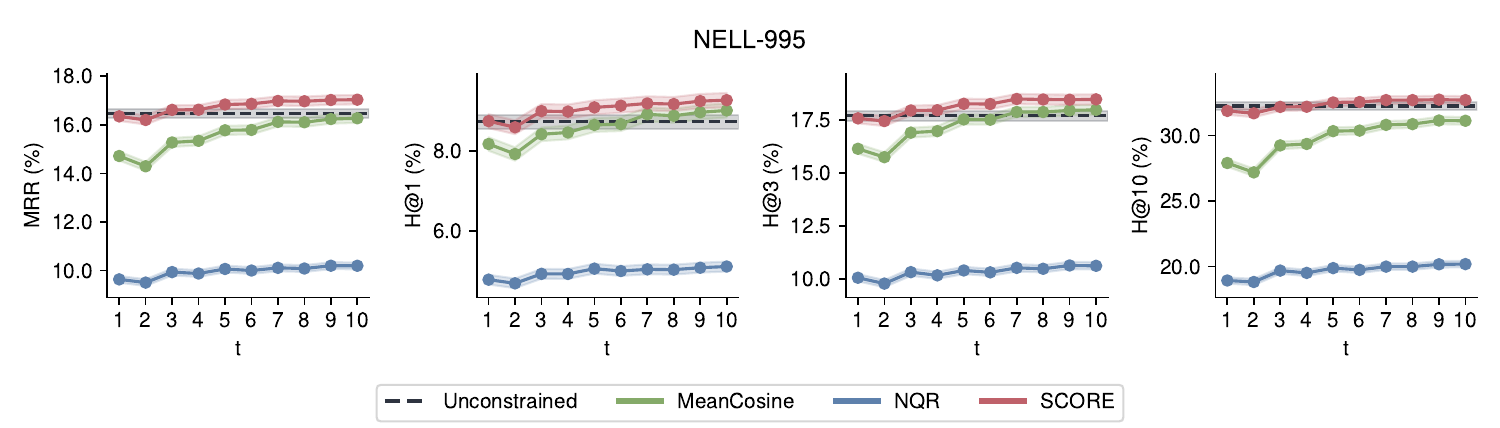}
    \end{subfigure}
    \caption{Additional metrics for SimCQA with similarity constraints on the \textbf{target} variable. Shaded areas indicate a 95\% confidence interval.}
    \label{fig:target_overtime_hits}
\end{figure*}

\begin{figure*}[t]
    \centering
    \begin{subfigure}{\textwidth}
        \centering
        \includegraphics[width=\textwidth]{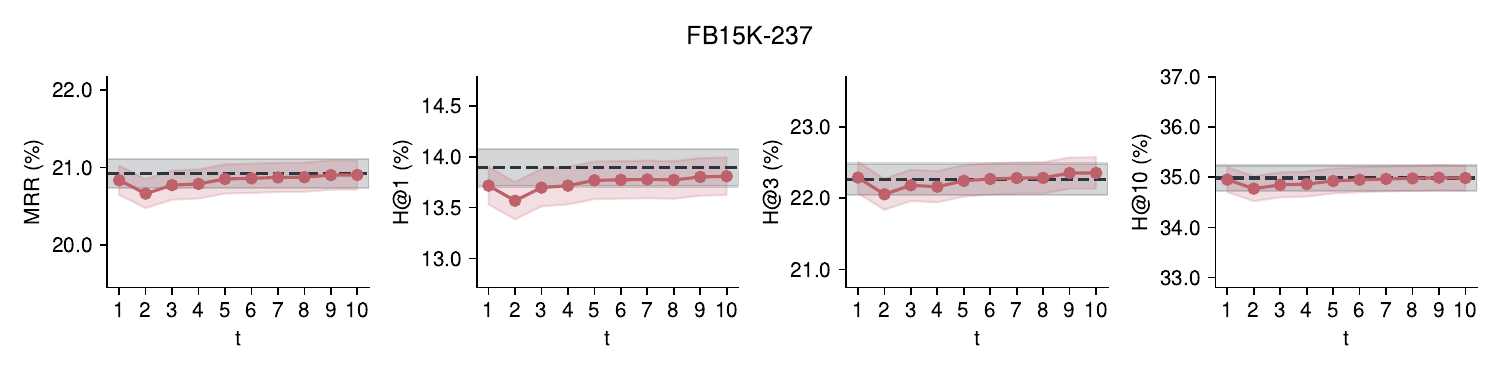}
        \includegraphics[width=\textwidth]{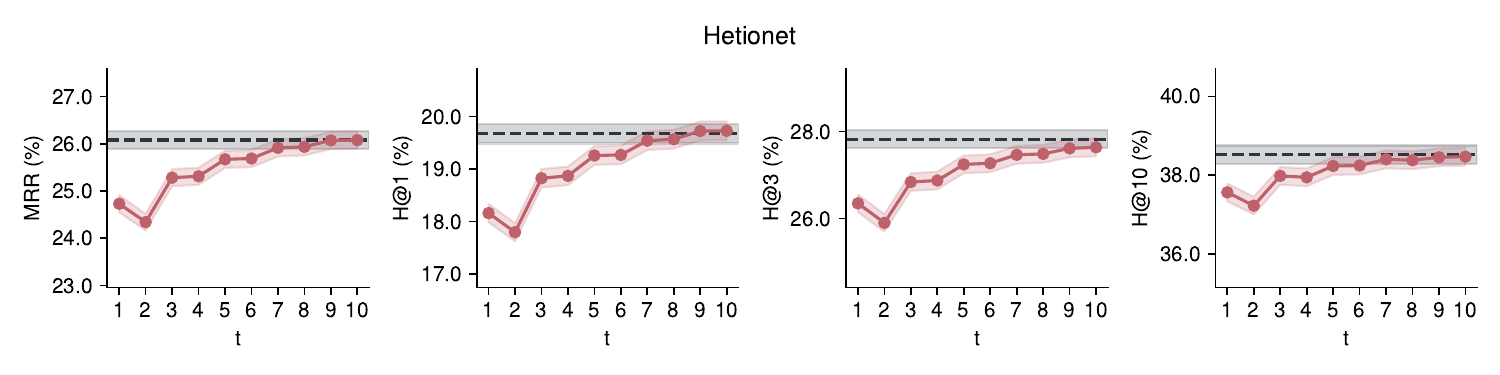}
        \includegraphics[width=\textwidth]{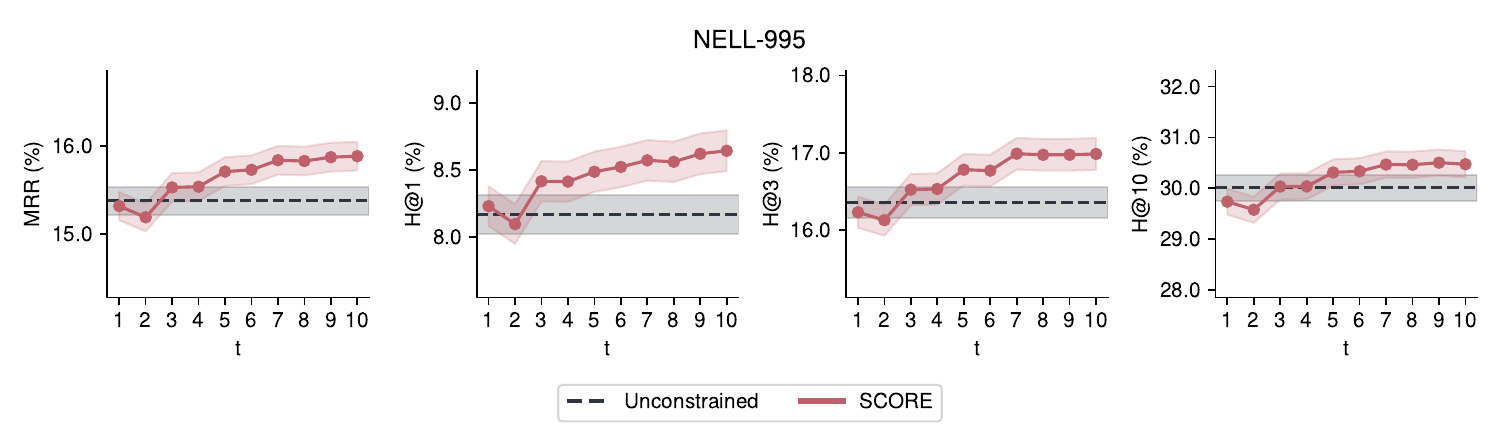}
    \end{subfigure}
    \caption{Additional metrics for SimCQA with similarity constraints on the \textbf{target and intermediate} variables. Shaded areas indicate a 95\% confidence interval.}
    \label{fig:full_overtime_hits}
\end{figure*}

\paragraph{Additional metrics.} We present additional results on more fine-grained ranking metrics, including H@3, in~\Cref{fig:target_overtime_hits} and~\Cref{fig:full_overtime_hits}.

\paragraph{Numeric results.} To facilitate direct comparison, \Cref{tab:numeric_fb15k237_target_comparison,tab:numeric_hetionet_target_comparison,tab:numeric_nell995_target_comparison,tab:numeric_fb15k237_full_comparison,tab:numeric_hetionet_full_comparison,tab:numeric_nell995_full_comparison} report the values shown in~\Cref{fig:target_overtime,fig:full_overtime} for preference sets of size $t\in\{1,5,10\}$.

\begin{table*}[t]
\centering
\caption{Numeric results for target-variable SimCQA on FB15k-237 at preference-set sizes $t\in\{1,5,10\}$.}
\label{tab:numeric_fb15k237_target_comparison}
\resizebox{\textwidth}{!}{%
\begin{tabular}{lccccccccccccccc}
\toprule
Method & \multicolumn{3}{c}{PA} & \multicolumn{3}{c}{MRR} & \multicolumn{3}{c}{H@1} & \multicolumn{3}{c}{H@10} & \multicolumn{3}{c}{NDCG@10} \\
\cmidrule(lr){2-4}\cmidrule(lr){5-7}\cmidrule(lr){8-10}\cmidrule(lr){11-13}\cmidrule(lr){14-16}
& $t=1$ & $t=5$ & $t=10$ & $t=1$ & $t=5$ & $t=10$ & $t=1$ & $t=5$ & $t=10$ & $t=1$ & $t=5$ & $t=10$ & $t=1$ & $t=5$ & $t=10$ \\
\midrule
Base CQA model & 54.7 & 54.7 & 54.7 & \textbf{24.2} & \textbf{24.2} & \textbf{24.2} & \textbf{16.3} & \textbf{16.3} & \textbf{16.3} & \textbf{40.0} & 40.0 & 40.0 & 33.4 & 33.4 & 33.4 \\
MeanCosine & 63.0 & 66.6 & 67.2 & 22.9 & 23.7 & 23.9 & 15.5 & 16.0 & 16.1 & 37.7 & 39.1 & 39.5 & 32.7 & 34.0 & 34.3 \\
NQR & \textbf{70.6} & \textbf{74.0} & \textbf{75.1} & 18.9 & 19.9 & 20.1 & 12.5 & 13.2 & 13.3 & 31.8 & 33.3 & 33.8 & 28.9 & 30.8 & 31.3 \\
SCORE & 58.6 & 60.2 & 60.3 & 24.0 & 24.1 & \textbf{24.2} & 16.1 & 16.2 & 16.2 & \textbf{40.0} & \textbf{40.1} & \textbf{40.1} & \textbf{34.1} & \textbf{34.4} & \textbf{34.4} \\
\bottomrule
\end{tabular}
}
\end{table*}

\begin{table*}[t]
\centering
\caption{Numeric results for target-variable SimCQA on Hetionet at preference-set sizes $t\in\{1,5,10\}$.}
\label{tab:numeric_hetionet_target_comparison}
\resizebox{\textwidth}{!}{%
\begin{tabular}{lccccccccccccccc}
\toprule
Method & \multicolumn{3}{c}{PA} & \multicolumn{3}{c}{MRR} & \multicolumn{3}{c}{H@1} & \multicolumn{3}{c}{H@10} & \multicolumn{3}{c}{NDCG@10} \\
\cmidrule(lr){2-4}\cmidrule(lr){5-7}\cmidrule(lr){8-10}\cmidrule(lr){11-13}\cmidrule(lr){14-16}
& $t=1$ & $t=5$ & $t=10$ & $t=1$ & $t=5$ & $t=10$ & $t=1$ & $t=5$ & $t=10$ & $t=1$ & $t=5$ & $t=10$ & $t=1$ & $t=5$ & $t=10$ \\
\midrule
Base CQA model & 51.7 & 51.7 & 51.7 & \textbf{32.5} & \textbf{32.5} & \textbf{32.5} & \textbf{25.1} & 25.1 & 25.1 & \textbf{47.2} & \textbf{47.2} & \textbf{47.2} & \textbf{41.5} & 41.5 & 41.5 \\
MeanCosine & 57.3 & 59.3 & 59.8 & 30.9 & 32.0 & 32.4 & 24.3 & \textbf{25.2} & \textbf{25.4} & 43.7 & 45.5 & 46.2 & 40.2 & 42.0 & 42.5 \\
NQR & \textbf{80.1} & \textbf{85.4} & \textbf{86.6} & 20.9 & 21.8 & 22.0 & 14.9 & 15.8 & 16.0 & 33.0 & 34.0 & 34.2 & 33.2 & 35.4 & 35.9 \\
SCORE & 56.9 & 58.0 & 58.1 & 30.7 & 31.8 & 32.4 & 23.0 & 24.4 & 25.1 & 45.9 & 46.6 & 46.9 & 41.1 & \textbf{42.4} & \textbf{42.9} \\
\bottomrule
\end{tabular}
}
\end{table*}

\begin{table*}[t]
\centering
\caption{Numeric results for target-variable SimCQA on NELL-995 at preference-set sizes $t\in\{1,5,10\}$.}
\label{tab:numeric_nell995_target_comparison}
\resizebox{\textwidth}{!}{%
\begin{tabular}{lccccccccccccccc}
\toprule
Method & \multicolumn{3}{c}{PA} & \multicolumn{3}{c}{MRR} & \multicolumn{3}{c}{H@1} & \multicolumn{3}{c}{H@10} & \multicolumn{3}{c}{NDCG@10} \\
\cmidrule(lr){2-4}\cmidrule(lr){5-7}\cmidrule(lr){8-10}\cmidrule(lr){11-13}\cmidrule(lr){14-16}
& $t=1$ & $t=5$ & $t=10$ & $t=1$ & $t=5$ & $t=10$ & $t=1$ & $t=5$ & $t=10$ & $t=1$ & $t=5$ & $t=10$ & $t=1$ & $t=5$ & $t=10$ \\
\midrule
Base CQA model & 51.7 & 51.7 & 51.7 & \textbf{16.5} & 16.5 & 16.5 & \textbf{8.7} & 8.7 & 8.7 & \textbf{32.2} & 32.2 & 32.2 & 26.7 & 26.7 & 26.7 \\
MeanCosine & 56.4 & 58.8 & 59.1 & 14.7 & 15.8 & 16.3 & 8.2 & 8.7 & 9.0 & 27.9 & 30.3 & 31.1 & 24.6 & 26.5 & 27.2 \\
NQR & \textbf{63.7} & \textbf{63.8} & \textbf{63.8} & 9.6 & 10.1 & 10.2 & 4.8 & 5.1 & 5.1 & 18.9 & 19.9 & 20.2 & 18.1 & 19.0 & 19.2 \\
SCORE & 53.9 & 54.3 & 54.3 & 16.3 & \textbf{16.8} & \textbf{17.0} & \textbf{8.7} & \textbf{9.1} & \textbf{9.3} & 31.9 & \textbf{32.5} & \textbf{32.7} & \textbf{26.8} & \textbf{27.7} & \textbf{27.8} \\
\bottomrule
\end{tabular}
}
\end{table*}

\begin{table*}[t]
\centering
\caption{Numeric results for general SimCQA on FB15k-237 at preference-set sizes $t\in\{1,5,10\}$.}
\label{tab:numeric_fb15k237_full_comparison}
\resizebox{\textwidth}{!}{%
\begin{tabular}{lccccccccccccccc}
\toprule
Method & \multicolumn{3}{c}{PA} & \multicolumn{3}{c}{MRR} & \multicolumn{3}{c}{H@1} & \multicolumn{3}{c}{H@10} & \multicolumn{3}{c}{NDCG@10} \\
\cmidrule(lr){2-4}\cmidrule(lr){5-7}\cmidrule(lr){8-10}\cmidrule(lr){11-13}\cmidrule(lr){14-16}
& $t=1$ & $t=5$ & $t=10$ & $t=1$ & $t=5$ & $t=10$ & $t=1$ & $t=5$ & $t=10$ & $t=1$ & $t=5$ & $t=10$ & $t=1$ & $t=5$ & $t=10$ \\
\midrule
Base CQA model & 53.3 & 53.3 & 53.3 & \textbf{20.9} & \textbf{20.9} & \textbf{20.9} & \textbf{13.9} & \textbf{13.9} & \textbf{13.9} & \textbf{35.0} & \textbf{35.0} & \textbf{35.0} & 31.7 & 31.7 & 31.7 \\
SCORE & 56.5 & 58.2 & 58.4 & 20.8 & \textbf{20.9} & \textbf{20.9} & 13.7 & 13.8 & 13.8 & \textbf{35.0} & 34.9 & \textbf{35.0} & \textbf{32.3} & \textbf{32.6} & \textbf{32.7} \\
MeanCosine & \textbf{58.6} & \textbf{61.1} & \textbf{61.4} & 17.3 & 17.8 & 17.9 & 11.4 & 11.7 & 11.8 & 28.7 & 29.7 & 30.0 & 26.1 & 27.0 & 27.2 \\
\bottomrule
\end{tabular}
}
\end{table*}

\begin{table*}[t]
\centering
\caption{Numeric results for general SimCQA on Hetionet at preference-set sizes $t\in\{1,5,10\}$.}
\label{tab:numeric_hetionet_full_comparison}
\resizebox{\textwidth}{!}{%
\begin{tabular}{lccccccccccccccc}
\toprule
Method & \multicolumn{3}{c}{PA} & \multicolumn{3}{c}{MRR} & \multicolumn{3}{c}{H@1} & \multicolumn{3}{c}{H@10} & \multicolumn{3}{c}{NDCG@10} \\
\cmidrule(lr){2-4}\cmidrule(lr){5-7}\cmidrule(lr){8-10}\cmidrule(lr){11-13}\cmidrule(lr){14-16}
& $t=1$ & $t=5$ & $t=10$ & $t=1$ & $t=5$ & $t=10$ & $t=1$ & $t=5$ & $t=10$ & $t=1$ & $t=5$ & $t=10$ & $t=1$ & $t=5$ & $t=10$ \\
\midrule
Base CQA model & 51.1 & 51.1 & 51.1 & \textbf{26.1} & \textbf{26.1} & \textbf{26.1} & \textbf{19.7} & \textbf{19.7} & \textbf{19.7} & \textbf{38.5} & \textbf{38.5} & \textbf{38.5} & \textbf{36.0} & 36.0 & 36.0 \\
SCORE & \textbf{55.3} & 56.6 & 56.8 & 24.7 & 25.7 & \textbf{26.1} & 18.2 & 19.3 & \textbf{19.7} & 37.6 & 38.2 & \textbf{38.5} & \textbf{36.0} & \textbf{37.3} & \textbf{37.7} \\
MeanCosine & 55.1 & \textbf{56.7} & \textbf{57.0} & 24.4 & 25.2 & 25.5 & 18.9 & 19.5 & \textbf{19.7} & 35.0 & 36.4 & 36.9 & 33.4 & 34.8 & 35.2 \\
\bottomrule
\end{tabular}
}
\end{table*}

\begin{table*}[t]
\centering
\caption{Numeric results for general SimCQA on NELL-995 at preference-set sizes $t\in\{1,5,10\}$.}
\label{tab:numeric_nell995_full_comparison}
\resizebox{\textwidth}{!}{%
\begin{tabular}{lccccccccccccccc}
\toprule
Method & \multicolumn{3}{c}{PA} & \multicolumn{3}{c}{MRR} & \multicolumn{3}{c}{H@1} & \multicolumn{3}{c}{H@10} & \multicolumn{3}{c}{NDCG@10} \\
\cmidrule(lr){2-4}\cmidrule(lr){5-7}\cmidrule(lr){8-10}\cmidrule(lr){11-13}\cmidrule(lr){14-16}
& $t=1$ & $t=5$ & $t=10$ & $t=1$ & $t=5$ & $t=10$ & $t=1$ & $t=5$ & $t=10$ & $t=1$ & $t=5$ & $t=10$ & $t=1$ & $t=5$ & $t=10$ \\
\midrule
Base CQA model & 51.4 & 51.4 & 51.4 & \textbf{15.4} & 15.4 & 15.4 & \textbf{8.2} & 8.2 & 8.2 & \textbf{30.0} & 30.0 & 30.0 & 26.5 & 26.5 & 26.5 \\
SCORE & 53.5 & 54.2 & 54.3 & 15.3 & \textbf{15.7} & \textbf{15.9} & \textbf{8.2} & \textbf{8.5} & \textbf{8.6} & 29.7 & \textbf{30.3} & \textbf{30.5} & \textbf{26.7} & \textbf{27.4} & \textbf{27.6} \\
MeanCosine & \textbf{55.1} & \textbf{57.0} & \textbf{57.2} & 12.8 & 13.6 & 14.0 & 7.0 & 7.4 & 7.7 & 24.1 & 26.0 & 26.6 & 22.7 & 24.1 & 24.6 \\
\bottomrule
\end{tabular}
}
\end{table*}

\begin{table*}[t]
    \caption{NDCG@10 results for \textbf{target-variable SimCQA}, averaged over preference sets of size $t=1,\ldots,10$.}
    \label{tab:detailed_target_results}
    \centering
    \resizebox{\textwidth}{!}{
\begin{tabular}{llllllllllllllll}
\toprule
Method & 1p & 2in & 2i & 2p & 2u & 3in & 3i & 3p & inp & ip & pin & pi & pni & up & Avg. \\
\midrule
\mc{16}{c}{FB15k237} \\
\midrule
Base CQA model & 43.90 & 23.34 & 47.40 & 28.75 & 34.37 & 37.14 & 50.54 & 27.16 & 21.48 & 34.32 & 20.05 & 41.32 & 9.87 & 27.98 & 31.97 \\
MeanCosine & 44.25 & 23.76 & 47.92 & 29.67 & 34.80 & 37.20 & 50.92 & 26.95 & \bf 22.81 & 34.29 & 20.38 & 40.53 & 10.33 & 28.34 & 32.30 \\
NQR & 41.61 & 21.13 & 46.13 & 25.24 & 32.33 & 33.82 & 49.43 & 21.30 & 19.48 & 27.79 & 16.92 & 36.12 & 9.25 & 25.95 & 29.04 \\
SCORE & \bf 44.43 & \bf 24.07 & \bf 48.06 & \bf 29.84 & \bf 34.91 & \bf 37.80 & \bf 51.38 & \bf 28.34 & 22.62 & \bf 35.61 & \bf 21.20 & \bf 42.03 & \bf 10.57 & \bf 28.73 & \bf 32.83 \\
\midrule
\mc{16}{c}{Hetionet} \\
\midrule
Base CQA model & 52.95 & 33.11 & 44.51 & 17.10 & 44.21 & 22.11 & 48.94 & 14.58 & 13.63 & 18.38 & 11.27 & 34.94 & 12.84 & 18.18 & 27.62 \\
MeanCosine & \bf 53.18 & 33.51 & 44.62 & 19.42 & \bf 44.62 & 23.56 & 47.97 & 13.56 & \bf 17.76 & 20.54 & \bf 14.35 & 31.66 & 13.35 & 18.76 & 28.35 \\
NQR & 45.24 & 24.65 & 39.93 & 14.65 & 35.57 & 19.20 & 44.64 & 11.75 & 13.20 & 13.67 & 10.96 & 26.11 & 11.23 & 16.13 & 23.35 \\
SCORE & 52.79 & \bf 33.59 & \bf 45.93 & \bf 20.69 & 44.31 & \bf 24.23 & \bf 50.38 & \bf 15.96 & 17.27 & \bf 23.53 & 14.23 & \bf 37.55 & \bf 14.41 & \bf 19.52 & \bf 29.60 \\
\midrule
\mc{16}{c}{NELL995} \\
\midrule
Base CQA model & 47.02 & 23.36 & \bf 41.79 & 29.45 & 33.53 & \bf 22.78 & 47.10 & 23.35 & 21.30 & 29.71 & 15.70 & 31.72 & 10.34 & 29.79 & 29.07 \\
MeanCosine & 46.98 & 22.99 & 41.74 & 28.85 & 32.81 & 21.74 & 47.36 & 22.64 & 22.23 & 27.67 & 14.60 & 28.97 & 10.17 & 29.35 & 28.43 \\
NQR & 37.55 & 14.81 & 33.77 & 20.28 & 26.46 & 13.23 & 37.64 & 14.15 & 15.04 & 21.62 & 7.94 & 23.04 & 6.25 & 21.19 & 20.93 \\
SCORE & \bf 47.58 & \bf 23.99 & 41.74 & \bf 30.43 & \bf 33.63 & 22.59 & \bf 47.57 & \bf 24.81 & \bf 22.98 & \bf 31.32 & \bf 16.54 & \bf 33.38 & \bf 11.02 & \bf 30.51 & \bf 29.86 \\
\bottomrule
\end{tabular}
}
\end{table*}

\begin{table*}[t]
    \caption{NDCG@10 results for \textbf{general SimCQA}, averaged over preference sets of size $t=1,\ldots,10$.}
    \label{tab:detailed_full_results}
    \centering
    \resizebox{\textwidth}{!}{
\begin{tabular}{llllllllllllllllllllllll}
\toprule
Method & 1ps0 & 2ins0 & 2is0 & 2ps0 & 2ps1 & 2us0 & 3ins0 & 3is0 & 3ps0 & 3ps1 & 3ps2 & inps0 & inps1 & ips0 & ips1 & pins0 & pins1 & pis0 & pis1 & pnis1 & ups0 & ups1 & Avg. \\
\midrule
\mc{24}{c}{FB15k237} \\
\midrule
Base CQA model & 43.90 & 23.34 & 47.40 & 29.29 & 28.75 & 34.37 & 37.14 & 50.54 & 28.60 & 27.72 & 27.16 & 22.15 & 21.48 & 40.29 & 34.32 & 21.48 & 20.05 & 36.97 & 41.32 & 9.87 & 30.34 & 27.98 & 31.11 \\
MeanCosine & 44.25 & 23.76 & 47.92 & 12.74 & 29.67 & 34.80 & 37.20 & 50.92 & 15.72 & 12.09 & 26.95 & 8.55 & \bf 22.81 & 14.34 & 34.29 & 9.23 & 20.38 & 20.19 & 40.53 & 10.33 & 13.41 & 28.34 & 25.38 \\
SCORE & \bf 44.43 & \bf 24.07 & \bf 48.06 & \bf 29.39 & \bf 29.84 & \bf 34.91 & \bf 37.80 & \bf 51.38 & \bf 29.00 & \bf 28.36 & \bf 28.34 & \bf 22.65 & 22.62 & \bf 40.42 & \bf 35.61 & \bf 22.36 & \bf 21.20 & \bf 38.12 & \bf 42.03 & \bf 10.57 & \bf 30.68 & \bf 28.73 & \bf 31.84 \\
\midrule
\mc{24}{c}{Hetionet} \\
\midrule
Base CQA model & 52.95 & 33.11 & 44.51 & 17.14 & 17.10 & 44.21 & 22.11 & 48.94 & 32.02 & 14.59 & 14.58 & 13.52 & 13.63 & 21.42 & 18.38 & 13.84 & 11.27 & 30.27 & 34.94 & 12.84 & 17.18 & 18.18 & 24.85 \\
MeanCosine & \bf 53.18 & 33.51 & 44.62 & 8.43 & 19.42 & \bf 44.62 & 23.56 & 47.97 & 28.39 & 9.01 & 13.56 & 7.96 & \bf 17.76 & 11.87 & 20.54 & 3.04 & \bf 14.35 & \bf 31.58 & 31.66 & 13.35 & 7.62 & 18.76 & 22.94 \\
SCORE & 52.79 & \bf 33.59 & \bf 45.93 & \bf 21.22 & \bf 20.69 & 44.31 & \bf 24.23 & \bf 50.38 & \bf 32.30 & \bf 20.55 & \bf 15.96 & \bf 16.62 & 17.27 & \bf 26.58 & \bf 23.53 & \bf 17.16 & 14.23 & 30.31 & \bf 37.55 & \bf 14.41 & \bf 18.59 & \bf 19.52 & \bf 27.17 \\
\midrule
\mc{24}{c}{NELL995} \\
\midrule
Base CQA model & 47.02 & 23.36 & \bf 41.79 & 30.98 & 29.45 & 33.53 & \bf 22.78 & 47.10 & 25.64 & 23.55 & 23.35 & 23.26 & 21.30 & 37.64 & 29.71 & 25.20 & 15.70 & 38.29 & 31.72 & 10.34 & 27.57 & 29.79 & 29.05 \\
MeanCosine & 46.98 & 22.99 & 41.74 & 16.10 & 28.85 & 32.81 & 21.74 & 47.36 & 21.69 & 13.37 & 22.64 & 11.89 & 22.23 & 22.86 & 27.67 & 14.21 & 14.60 & \bf 40.49 & 28.97 & 10.17 & 16.04 & 29.35 & 25.21 \\
SCORE & \bf 47.58 & \bf 23.99 & 41.74 & \bf 31.45 & \bf 30.43 & \bf 33.63 & 22.59 & \bf 47.57 & \bf 26.78 & \bf 25.07 & \bf 24.81 & \bf 24.52 & \bf 22.98 & \bf 39.30 & \bf 31.32 & \bf 26.21 & \bf 16.54 & 40.30 & \bf 33.38 & \bf 11.02 & \bf 27.63 & \bf 30.51 & \bf 29.97 \\
\bottomrule
\end{tabular}
}
\end{table*}

\begin{figure*}[t]
    \centering
    \begin{subfigure}{\textwidth}
        \centering
        \includegraphics[width=\textwidth]{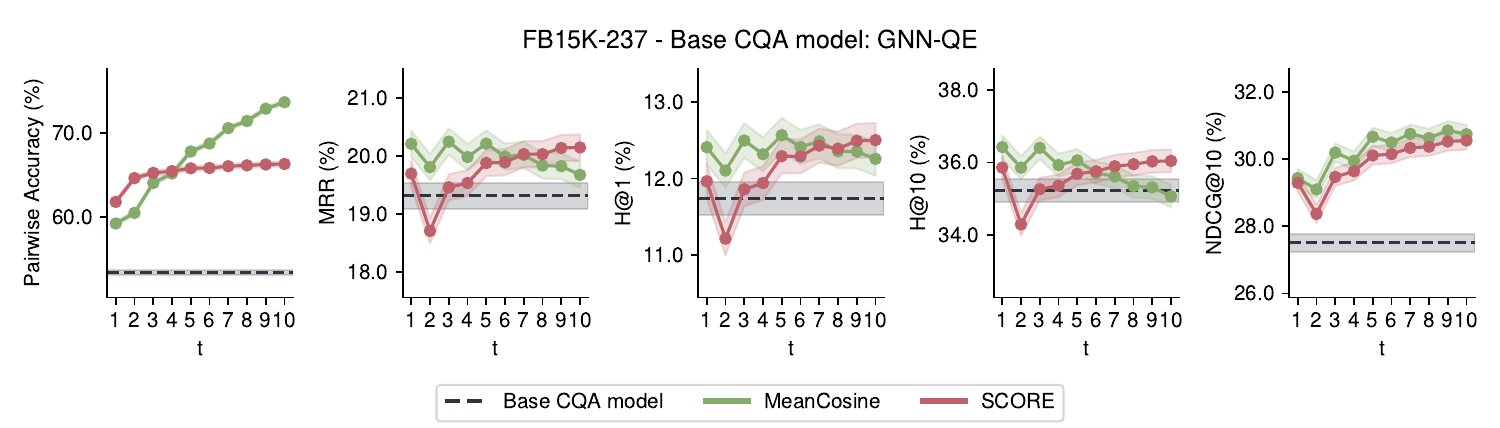}
        \includegraphics[width=\textwidth]{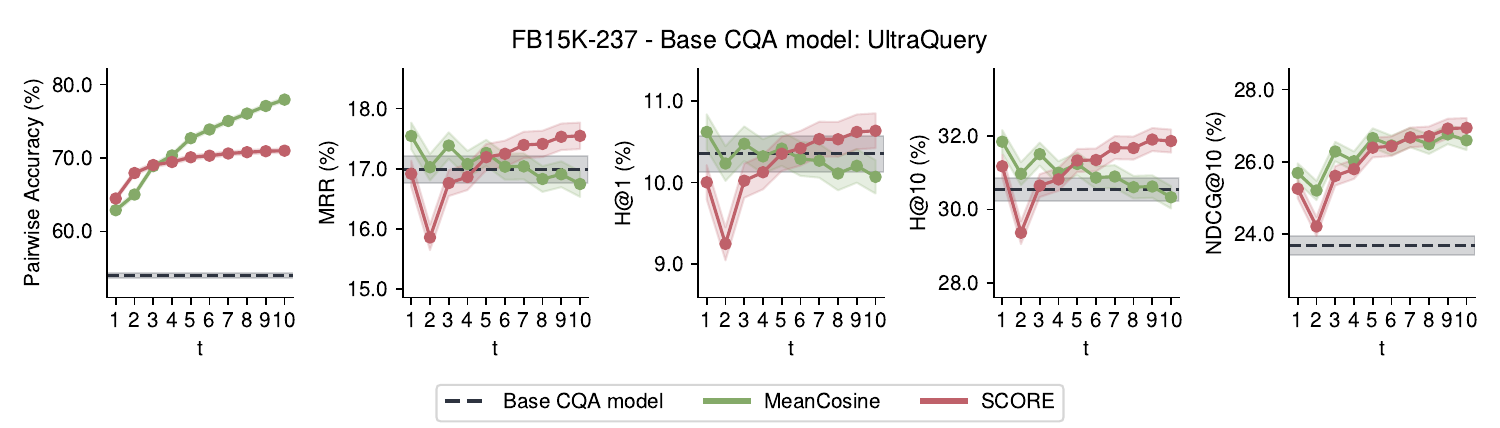}
        \includegraphics[width=\textwidth]{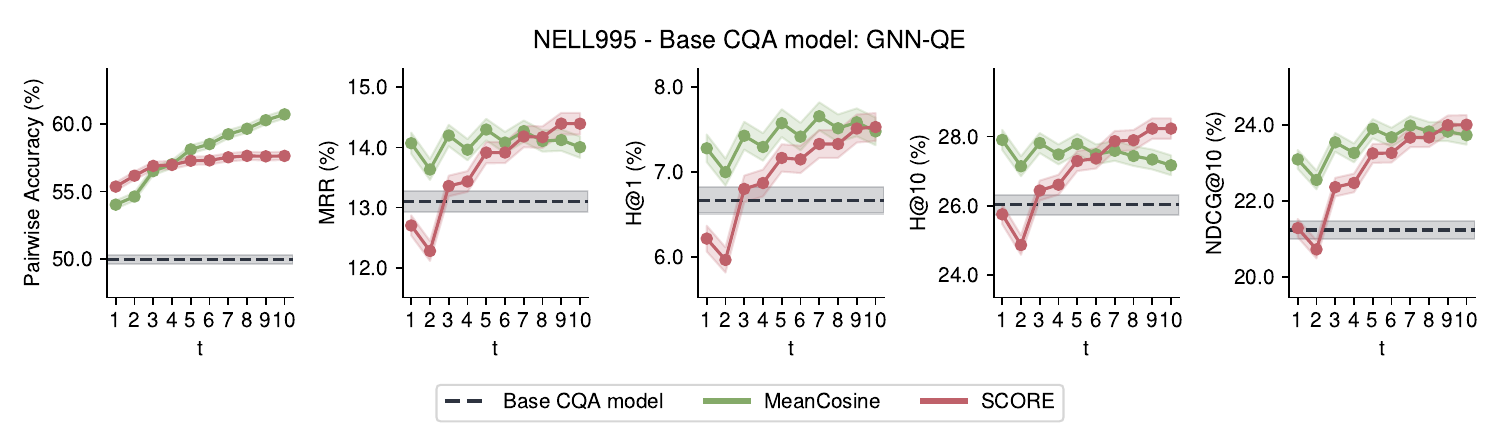}
        \includegraphics[width=\textwidth]{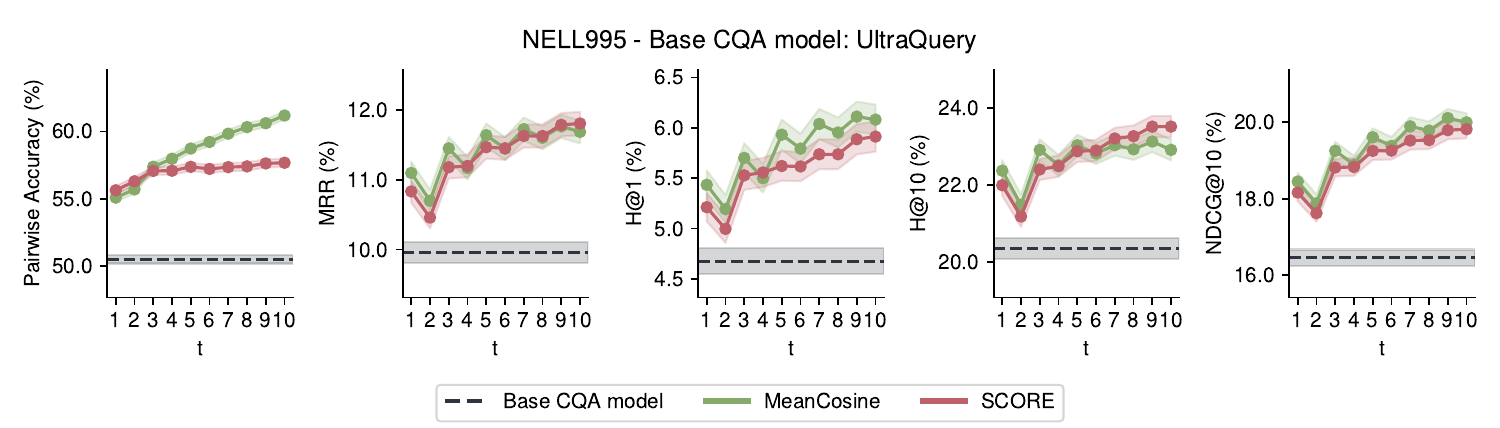}
    \end{subfigure}
    \caption{SimCQA results for similarity constraints on the \textbf{target} variable, using different base CQA models: GNN-QE~\citep{zhu_neural-symbolic_2022} and UltraQuery~\citep{galkin_foundation_2024}. Shaded areas indicate a 95\% confidence interval.}
    \label{fig:target_overtime_gnnqe_ultra}
\end{figure*}

\begin{figure*}[t]
    \centering
    \begin{subfigure}{\textwidth}
        \centering
        \includegraphics[width=\textwidth]{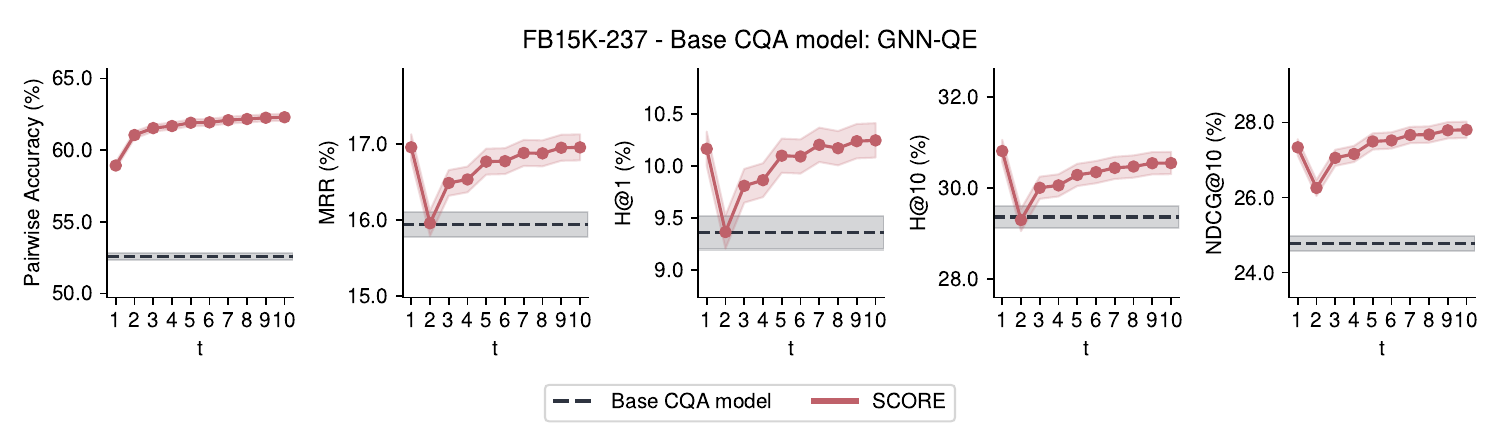}
        \includegraphics[width=\textwidth]{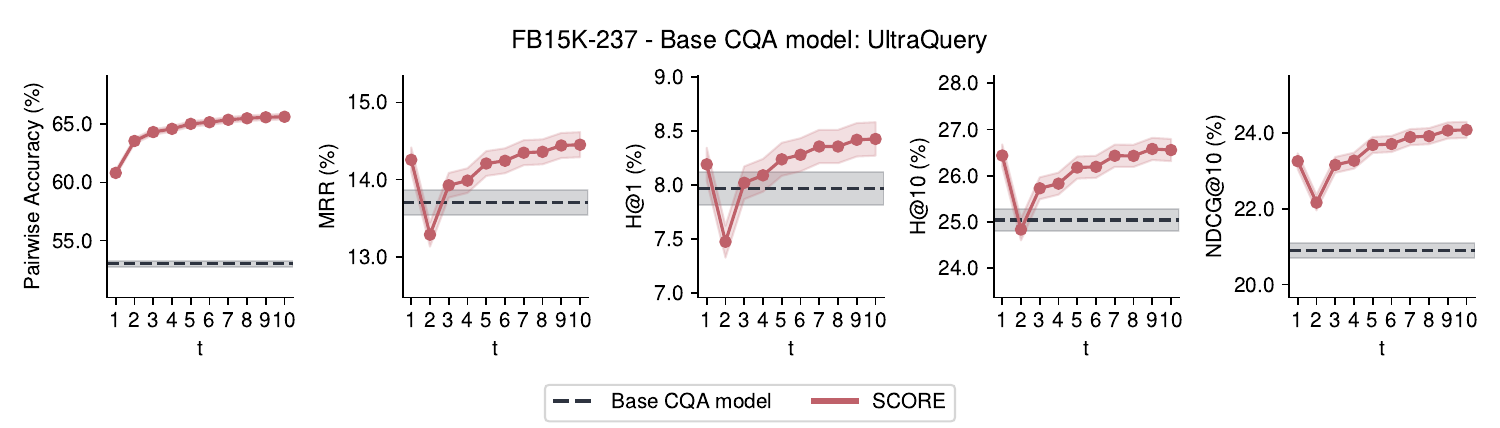}
        \includegraphics[width=\textwidth]{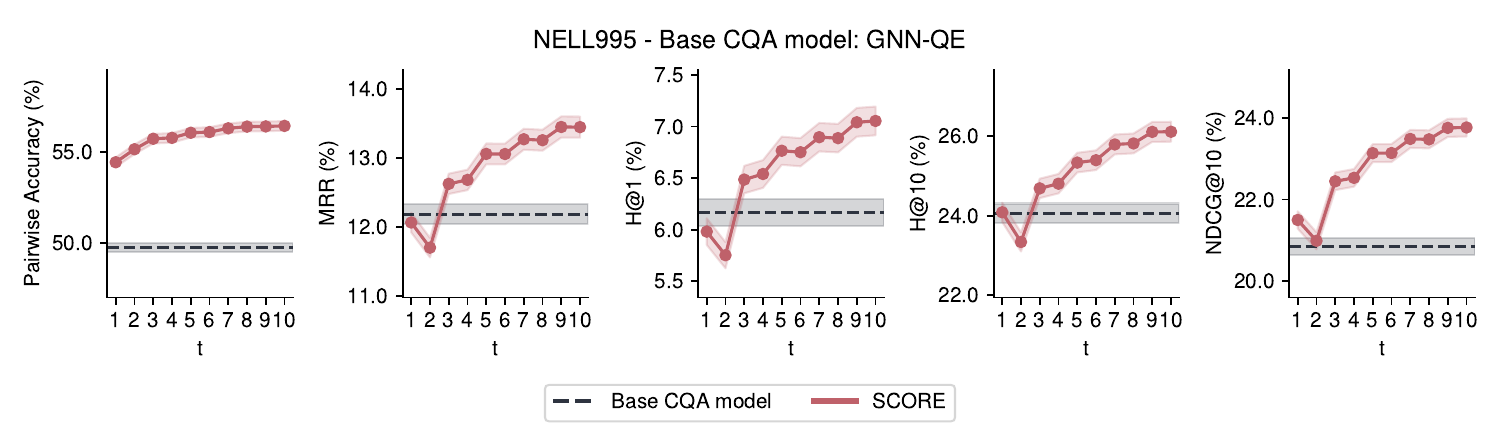}
        \includegraphics[width=\textwidth]{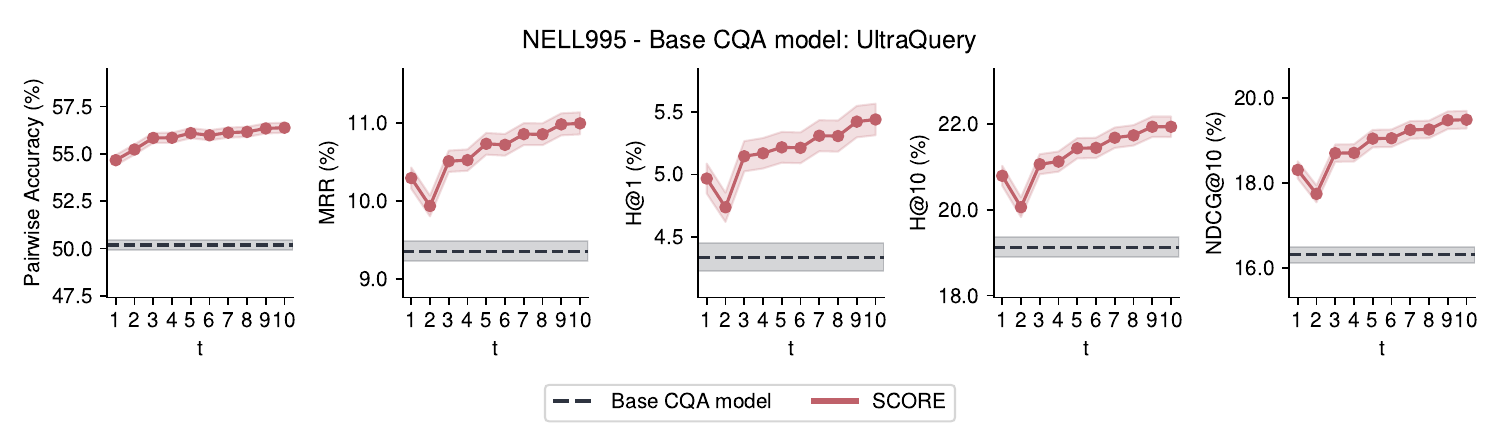}
    \end{subfigure}
    \caption{SimCQA results for similarity constraints on \textbf{target and intermediate} variables, using different base CQA models: GNN-QE~\citep{zhu_neural-symbolic_2022} and UltraQuery~\citep{galkin_foundation_2024}. Shaded areas indicate a 95\% confidence interval.}
    \label{fig:full_overtime_gnnqe_ultra}
\end{figure*}

\paragraph{Alternative base CQA models.} We present further results on both target-variable and general SimCQA when replacing the base CQA models with GNN-QE~\citep{zhu_neural-symbolic_2022} and UltraQuery~\citep{galkin_foundation_2024} in~\Cref{fig:target_overtime_gnnqe_ultra} and~\Cref{fig:full_overtime_gnnqe_ultra}. We observe the same qualitative behavior previously reported with QTO: incorporating similarity constraints with SCORE consistently improve preference satisfaction while preserving or improving global ranking performance, even under changes of the base CQA models.

\begin{table*}[t]
    \caption{Pairwise accuracy (PA) and MRR per query type for \textbf{target-variable SimCQA} on FB15k-237, averaged over preference sets of size $t=1,\ldots,10$. $\Delta$ denotes the difference between SCORE and the base CQA model.}
    \label{tab:per_type_pa_mrr_fb15k237}
    \centering
    \resizebox{\textwidth}{!}{
\begin{tabular}{llllllllllllllll}
\toprule
Method & 1p & 2in & 2i & 2p & 2u & 3in & 3i & 3p & inp & ip & pin & pi & pni & up & Avg. \\
\midrule
\mc{16}{c}{Pairwise accuracy} \\
\midrule
Base CQA model & 58.15 & 56.03 & 54.72 & 55.59 & 57.70 & 52.98 & 54.22 & 53.01 & 54.65 & 55.55 & 53.62 & 55.50 & 57.39 & 53.97 & 55.22 \\
MeanCosine & 63.36 & 67.45 & 58.95 & 69.59 & 66.01 & 61.64 & 57.60 & 70.98 & 69.95 & 68.18 & 71.60 & 64.55 & 69.81 & 67.07 & 66.20 \\
NQR & 79.45 & 75.37 & 69.22 & 73.89 & 74.47 & 70.45 & 67.28 & 72.66 & 72.37 & 75.71 & 73.72 & 72.71 & 75.79 & 71.78 & 73.20 \\
SCORE & 62.82 & 62.44 & 58.06 & 61.04 & 62.12 & 58.92 & 57.28 & 58.23 & 59.58 & 60.75 & 60.42 & 59.15 & 62.34 & 58.75 & 60.14 \\
$\Delta$ & +4.67 & +6.41 & +3.34 & +5.46 & +4.42 & +5.94 & +3.06 & +5.21 & +4.92 & +5.20 & +6.80 & +3.65 & +4.94 & +4.78 & +4.91 \\
\midrule
\mc{16}{c}{MRR} \\
\midrule
Base CQA model & 35.01 & 16.19 & 34.13 & 21.09 & 27.11 & 22.99 & 34.80 & 19.49 & 14.51 & 25.95 & 13.96 & 29.79 & 6.94 & 20.07 & 23.00 \\
MeanCosine & 34.75 & 15.87 & 34.06 & 20.67 & 26.91 & 22.69 & 34.78 & 17.63 & 14.32 & 24.15 & 12.99 & 28.27 & 6.81 & 19.51 & 22.39 \\
NQR & 29.34 & 13.51 & 30.52 & 16.38 & 23.78 & 19.35 & 31.45 & 12.91 & 11.51 & 17.90 & 10.18 & 23.45 & 5.86 & 16.69 & 18.77 \\
SCORE & 34.53 & 16.06 & 34.13 & 21.13 & 26.86 & 22.93 & 35.11 & 19.55 & 14.76 & 25.97 & 14.16 & 29.93 & 7.06 & 20.04 & 23.02 \\
$\Delta$ & $-$0.48 & $-$0.13 & +0.00 & +0.04 & $-$0.25 & $-$0.06 & +0.31 & +0.06 & +0.25 & +0.02 & +0.20 & +0.14 & +0.12 & $-$0.03 & +0.01 \\
\bottomrule
\end{tabular}
}
\end{table*}

\subsection{Robustness to noisy preferences}
\label{app:noise}

\begin{figure*}[t]
    \centering
    \includegraphics[width=\textwidth]{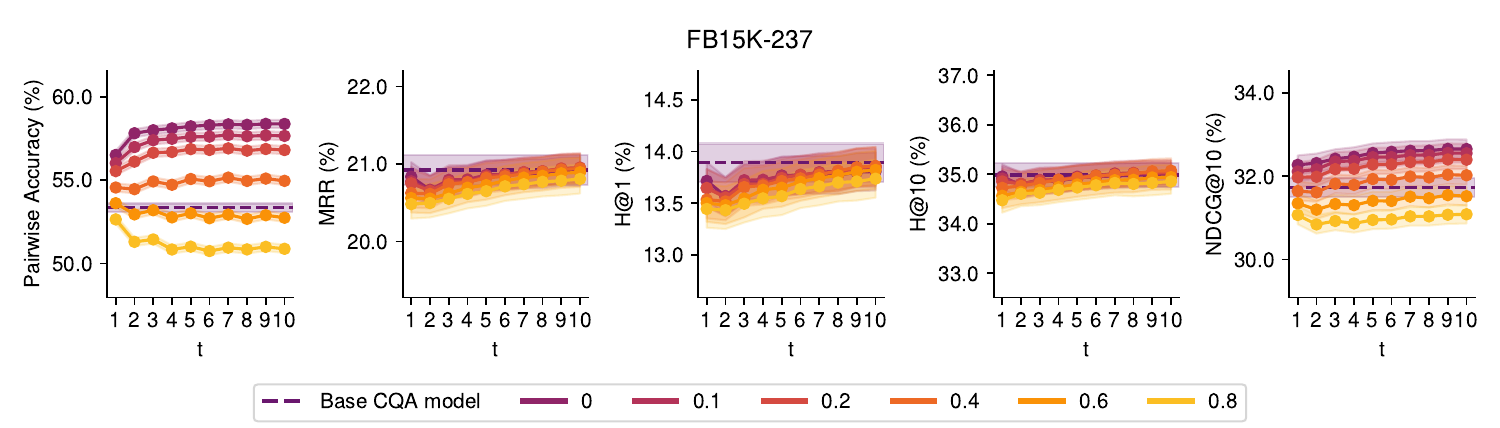}
    \caption{SCORE for general SimCQA on FB15k-237 when a fraction $k$ of the labels in each preference set is flipped. Shaded areas indicate a 95\% confidence interval.}
    \label{fig:noise_robustness}
\end{figure*}

To simulate imperfect feedback, we independently flip the label of a fraction $k\in\{0, 0.1, 0.2, 0.4, 0.6, 0.8\}$ of the entities in each preference set, so that preferred entities are treated as non-preferred and vice versa. As shown in~\Cref{fig:noise_robustness}, noise mainly affects pairwise accuracy: at $t=10$, it decreases from 58.4 without noise to 55.0 at $k=0.4$, still above the base CQA model (53.3), and falls below it only when feedback becomes predominantly misleading (50.9 at $k=0.8$). In contrast, MRR changes only slightly, from 20.9 to 20.8 at $k=0.8$. SCORE therefore tolerates moderate noise while largely preserving global answer quality. Regarding imbalanced preference sets, SCORE averages logit similarities within $\prefpos$ and $\prefneg$ separately, so that a larger set does not dominate the update merely by containing more examples.

\subsection{Alignment between embeddings and preferences}
\label{app:alignment}

SCORE approximates similarity constraints with entity embeddings from the base CQA model, so its effectiveness depends on how well the embedding space reflects a preference. To measure this, for each preference set we build an average-linkage hierarchical clustering of its entities $\prefpos\cup\prefneg$ using the cosine similarity of the ComplEx embeddings of QTO, and identify the smallest cluster $S$ containing all preferred entities $\prefpos$. We define its precision as $\vert\prefpos\vert/\vert S\vert$, which equals 1 if the embedding space separates preferred from non-preferred entities perfectly. We compute it for the full preference sets of target-variable SimCQA on the FB15k-237 test set, and average it per query.

Precision ranges from 0.16 to 1.0, and is perfect for only 1.3\% of the queries. For 29\% of the queries, precision is below 0.5, meaning that $S$ contains more non-preferred than preferred entities. This occurs most often for 3i (35\%) and 2i (34\%) queries, which are also the query types with the smallest gains in pairwise accuracy (+3.06 and +3.34, respectively, see~\Cref{tab:per_type_pa_mrr_fb15k237}). Precision is positively associated with the pairwise accuracy of SCORE, with a Spearman correlation of 0.20 overall and between 0.15 and 0.29 for every query type (all significant at $p<0.05$ after Benjamini--Hochberg correction). By contrast, its correlation with MRR is close to zero overall ($-0.03$), and significant only for 1p, 2i, and 3i queries. When the embedding space is poorly aligned with a preference, SCORE is thus less able to capture it, while global ranking quality is not systematically affected.

\subsection{Multiple simultaneous constraints}
\label{app:multiple_constraints}

\begin{table}[t]
\centering
\small
\caption{Results with similarity constraints on two variables of 2p and 3p queries (FB15k-237), averaged over preference sets of size $t=1,\ldots,10$.}
\label{tab:multiple_constraints}
\begin{tabular}{lrrrrrr}
\toprule
& \mc{2}{c}{PA} & \mc{2}{c}{MRR} & \mc{2}{c}{NDCG@10} \\
\cmidrule(lr){2-3}\cmidrule(lr){4-5}\cmidrule(lr){6-7}
Method & 2p & 3p & 2p & 3p & 2p & 3p \\
\midrule
Base CQA model & 52.54 & 49.16 & \bf 18.07 & 11.54 & 26.76 & 25.64 \\
SCORE & \bf 58.17 & \bf 54.48 & 17.96 & \bf 11.58 & \bf 27.88 & \bf 26.23 \\
\bottomrule
\end{tabular}
\end{table}

We evaluate SCORE with similarity constraints applied simultaneously to the first two variables of 2p and 3p queries on FB15k-237 (648 and 504 test queries, respectively). Following the protocol in~\Cref{app:datasets}, we first cluster the bindings of the first variable, and then cluster the bindings of the second variable that are reachable from the preferred cluster of the first. Preferred answers are those reached only through bindings that satisfy both constraints. In each interaction step, one example is added to the preference set of each variable, and SCORE applies each update as the corresponding fuzzy vector is computed. \Cref{tab:multiple_constraints} reports the results. As in single-constraint settings, SCORE substantially improves pairwise accuracy and NDCG@10, while MRR remains close to the base CQA model.


\ifincludechecklist
\clearpage
\input{checklist.tex}
\fi

\end{document}